\documentclass[sigconf,screen,nonacm]{acmart}

\AtBeginDocument{%
  }

\usepackage[normalem]{ulem}
\usepackage{subcaption}
\usepackage{makecell}
\usepackage{microtype}

\renewcommand{\bibliofont}{\footnotesize}

\begin{document}

\title{\texttt{slang.gr} as a Large-Scale Crowdsourced Resource for Non-Standard Greek}
\date{}

\author{Panagiotis Papadakos}
\email{papadako@ics.forth.gr}

\affiliation{%
\institution{ICS-FORTH, Heraklion}
\country{Greece}
}

\author{Katerina  Papantoniou}
\email{papanton@ics.forth.gr}
\affiliation{%
\institution{ICS-FORTH, Heraklion}
\country{Greece}
}

\author{Dimitris Plexousakis}
\email{dp@ics.forth.gr}
\affiliation{%
  \institution{ICS-FORTH, Heraklion}
\country{Greece}
}

\additionalaffiliation{%
\institution{University of Crete, Computer Science Department, Heraklion, Greece}
}

\renewcommand{\shortauthors}{Papadakos et al.}

\begin{abstract}
Slang is a central component of everyday language, reflecting linguistic creativity, social identity, and cultural change, yet its dynamic and non-standard nature makes it difficult to model computationally. We present the first large-scale computational study of \texttt{slang.gr}, a crowdsourced lexicon of Greek non-standard language, combining lexical content, user-generated tags, and interaction data.
To enable the systematic analysis, we map noisy folksonomic tags to a structured multi-layer taxonomy capturing both semantic categories and sociolinguistic metadata. Using this representation, we analyze the linguistic structure of Greek slang and the behavior of its contributor community. We find that slang is strongly centered on person-related and evaluative language, exhibits high morphological creativity, and is shaped by highly skewed participation with short user lifespans and overlapping communities.
Building on these signals, we introduce a community-based confidence score for definitions that integrates user roles, interaction patterns, and moderation signals. Our results show that taxonomy-based representations improve interpretability while retaining meaningful aspects of behavioral structure, enabling a more structured and interpretable analysis of confidence signals.
Overall, this work establishes \texttt{slang.gr} as a computational resource for non-standard Greek and provides a foundation for sociolinguistic NLP, bias analysis, and the study of informal language in LLMs.
\end{abstract}



\maketitle

\begin{center}
\small\itshape
Preprint of a paper accepted for publication in the Proceedings of the 14th
EETN Conference on Artificial Intelligence (SETN 2026).
\end{center}

\section{Introduction}
Slang and non-standard language play a central role in everyday communication, both oral and written.
Slang is \textit{an ever-changing set of colloquial words and phrases that speakers use to establish or reinforce social identity or cohesiveness within a group or with a trend or fashion in society at large}~\cite{eble1996slang}. Another definition characterizes slang as an informal lexical phenomenon associated with group identification and marked deviation from standard usage, often signaling familiarity and considered inappropriate in formal contexts~\cite{dumas1978slang}. Sociologically, it functions as a marker of group identity and cohesion, enabling both insider bonding and outsider exclusion, and thus can be both playful and aggressive~\cite{mattiello2008introduction}.

As a form of non-standard language, slang offers a valuable window into linguistic creativity, social meaning, and variation. It is also of particular relevance for modern AI systems, especially large language models (LLMs). Slang captures rapidly evolving meanings, informal usage patterns, and socially embedded interpretations that are often underrepresented in standard training corpora. This poses significant challenges for LLMs, since meaning is often highly context-dependent and polysemous. The same expression may convey different meanings depending on speaker intent and social context, and non-standard orthography and creative word formation reduce lexical coverage and increase sparsity. Moreover, slang frequently relies on implicit cultural knowledge, irony, and pragmatic cues that are difficult to infer from text alone. As a result, LLMs may misinterpret meaning, fail to capture nuance, or overgeneralize offensive or biased usage. At the same time, slang can serve as a natural testbed for evaluating model robustness, the ability to capture variation, and the capacity to represent creativity, bias, and socially grounded meaning in non-standard language.

Despite its importance, slang remains difficult to model computationally due to its dynamic, context-dependent, and figurative nature. At the same time, the rise of user-generated content has facilitated the documentation of such phenomena. A notable example is \texttt{slang.gr}\footnote{\url{https://slang.gr}}, a crowdsourced lexicon of Greek slang. Despite its richness, the dataset exhibits noise and heterogeneity, including orthographic variation, overlapping and redundant tags, and inconsistent categorization. As a result, such resources are not immediately suitable for computational use and require normalization and alignment with a structured taxonomy that captures both semantic and metadata dimensions.

This work makes three main contributions:
(i) a structured multi-layer taxonomy that maps noisy user-generated tags into semantic and sociolinguistic dimensions,
(ii) a large-scale analysis of Greek slang covering both linguistic structure and community dynamics,
and (iii) a community-based confidence metric that integrates user roles, interaction patterns, and moderation signals. All resources and code are publicly available\footnote{Code and resources at \url{https://gitlab.isl.ics.forth.gr/slang/slang.gr}}.

We first describe the dataset and its folksonomic tag structure, then propose a taxonomy inspired by the Oxford Dictionary of Modern Slang~\cite{ayto2010oxford}, extended with metadata dimensions capturing linguistic form, usage context, and pragmatic stance. Tag mapping is performed using a hybrid approach combining LLM-based assignments with manual curation. Using this framework, we analyze the semantic and sociolinguistic organization of \texttt{slang.gr} and examine the contributor user roles, collaboration patterns, temporal dynamics, community structure and interactions, and contribution inequality and definition confidence.

The paper is organized as follows: Section~\ref{sec:rw} discusses related work, while Section~\ref{sec:dataset} describes the dataset and user-assigned tags. Section~\ref{sec:taxonomy} presents the proposed Oxford-inspired taxonomy and details the tag mapping process and Section~\ref{sec:community} analyzes the \texttt{slang.gr} community signals. Finally, Section~\ref{sec:conc} concludes the paper. 

Overall, this work provides the first large-scale computational analysis of \texttt{slang.gr} as a sociolinguistic and community-driven resource. By combining lexical, tagging, and interaction data with a structured taxonomy, we enable more interpretable analyses of linguistic variation and user behavior, and derive signals for assessing slang definition confidence.

\section{Related Work}
\label{sec:rw}

The computational analysis of internet slang  in~\cite{kulkarni2017tfw} is based on the Urban Dictionary\footnote{\url{https://www.urbandictionary.com}} and showcases that slang exhibits distinctive phonological patterns (e.g., increased use of fricatives and affricates), extra-grammatical morphology (e.g., non-standard prefixes), and a higher proportion of nouns compared to standard English. At the social level, topics such as sex and drugs dominate, and slang displays stronger gender, sexual, and religious biases than standard language. The work in~\cite{keidar2022slangvolution} shows that slang terms exhibit stable meanings but frequency shifts over time on Twitter.
Other works have addressed slang detection~\cite{Pei2019slang,aloraini2026slangtrack}, generation~\cite{sun2021generation}, and interpretation~\cite{sun2022interpretation}, rather than the linguistic properties of slang.

In the context of LLMs, recent work highlights the difficulty of modeling culturally and linguistically marked language that leads to systematic performance disparities across languages and cultural settings. For example,~\cite{wuraola2024understanding} demonstrate significant gaps in slang understanding across cultural contexts, with models performing better on Western than non-Western data, which is also common in other tasks such as deception detection ~\cite{papantoniou2025deception}. Similarly,~\cite{pedersen2025danish} and~\cite{attia2026llmarab} show that LLMs struggle with culture-specific figurative language in Danish and Arabic, respectively. 

Despite this growing body of work, computational studies of slang remain largely focused on English. To the best of our knowledge, no comparable large-scale analysis exists for non-standard Greek. The work in~\cite{xydopoulos2010slang} documents \texttt{slang.gr} from a lexicographic perspective, while~\cite{cyslang2019} presents Lex-Cy for Cypriot Greek slang. Additional resources include IDION~\cite{markantonatou2019idion}, a database of Modern Greek multiword expressions, and offensive language lexicons such as HURTLEX(el)~\cite{Stamou2022GOL}, which provides a multidimensional categorization of 737 offensive terms in Greek. However, these resources do not offer a unified computational framework for analyzing slang across semantic, sociolinguistic, and pragmatic dimensions.

Moreover, despite the extensive work on slang in English, we are not aware of any comprehensive computational taxonomy that systematically organizes slang. This work addresses these gaps by providing the first large-scale computational analysis of Greek slang, introducing a unified multi-layer taxonomy that organizes slang according to meaning, linguistic form, usage context, and social function. By leveraging this taxonomy, we transform \texttt{slang.gr} into a structured resource that supports the systematic study of linguistic properties, user community dynamics, and signals for assessing slang definition confidence, enabling cross-lingual comparison and improving the handling of non-standard language in NLP systems and LLMs.

\section{slang.gr Dataset Description}
\label{sec:dataset}

The dataset used in this study was scraped from \texttt{slang.gr}\footnote{According to the \texttt{slang.gr} terms of use, the data may be downloaded and analyzed for scientific research purposes.}, a large, collaboratively curated online repository of non-standard Greek language that was first documented in~\cite{xydopoulos2010slang}.
Although presented as a slang lexicon, \texttt{slang.gr} covers a broader range of non-standard and colloquial expressions. It is a crowdsourced repository of non-standard Greek, with slang as a core component.

The platform, which was launched in 2006, functions as a community-driven lexicon where users contribute entries, senses, and tags that reflect contemporary and historical non-standard language usage. As such, the dataset captures not only lexical items, but also the social, cultural, and pragmatic contexts in which slang is produced and interpreted. Each entry in the dataset is organized around a \textit{lemma}, corresponding to a slang term or expression. A lemma may be associated with one or more \textit{senses}, each representing a distinct definition or usage of the term. These senses are typically accompanied by \textit{usage examples}, which provide contextualized instances of the slang expression in natural language. This structure allows for a fine-grained representation of meaning, capturing both polysemy and contextual variation.
Senses are accompanied by \textit{tags} that are noisy, heterogeneous, and partially redundant. We normalize and canonicalize them by resolving orthographic variation and grouping semantically equivalent tags. 
Finally, each sense can be accompanied by user \textit{comments}.

The dataset contains 28,384 sense-level entries corresponding to 24,555 unique lemmas (1.156 senses per lemma). 
The lemmas exhibiting the highest number of distinct senses are \textit{παντόφλα} (14), \textit{αυγό} (11), \textit{τάπα} (10), \textit{μ*****ς} (10), \textit{κλέφτης} (9), and \textit{L.A.} (9). The lemmas are annotated with 649 unique normalized tags leading to 94,976 tag occurrences. During preprocessing, 4,312 tag normalization operations were performed to resolve orthographic and semantic variation. The total number of comments is 101,957.

In addition, the \texttt{slang.gr} data offers rich behavioural data about its contributors, allowing the study of its 17,942 registered users who have shaped the lexicon over two decades, examining user roles, collaboration networks, engagement patterns, temporal dynamics, and contribution inequality and confidence (Section \ref{sec:community}).

\subsection{Category-Level Tag Distribution}
\label{sec:categories}

A subset of the crowdsourced tags is explicitly organized as \textit{Category: Value}, where the category denotes a higher-level dimension and the value specifies a particular attribute. Table~\ref{tab:category-distribution} summarizes these categories. More than 75\% of tag occurrences are \textit{Uncategorized}, while \textit{Grammar}, \textit{Affective Load}, and \textit{Foreign Influences} account for most structured annotations. This imbalance motivates the construction of a structured taxonomy.
During normalization, the uncategorized tag ``Foreign Influences''
was assigned to the unspecified value of the corresponding category.

\begin{table}[ht]
\centering
\footnotesize
\begin{tabular}{lrr}
\hline
\textbf{Category} & \textbf{Occurrences} & \textbf{Unique Tags} \\
\hline
Uncategorized & 72,144 & 460 \\
Grammar & 9,967 & 111 \\
Affective Load & 6,712 & 24 \\
Foreign Influences & 4,541 & 20 \\
Region & 1,067 & 19 \\
Temporal Dimension & 545 & 15 \\
\hline
\textbf{Total} & \textbf{94,976} & \textbf{649} \\
\hline
\end{tabular}
\caption{Distribution of tags across dataset categories.}
\label{tab:category-distribution}
\end{table}

\subsubsection{Grammatical (Γραμματική) mapping.}

Grammar tags encode part-of-speech (e.g., \textit{noun}), word formation (e.g., \textit{compounding}), and rhetorical or stylistic devices (e.g., \textit{metaphor}). Their distribution shown in Table~\ref{tab:grammar} highlights the importance of morphological creativity and figurative language in Greek slang, with processes such as derivation, blending, and semantic shift occurring frequently. 

\begin{table}[ht]
\centering
\footnotesize
\begin{tabular}{lr@{\hspace{1.5em}}lr}
\hline
\textbf{Grammatical Category} & \textbf{Freq.} & \textbf{Grammatical Category} & \textbf{Freq.} \\
\hline
Word formation & 2313 & Part of speech -- adverb & 162 \\
Figure of speech -- wordplay & 1649 & Part of speech -- interjection & 150 \\
Part of speech -- noun & 1537 & Figure of speech -- simile & 146 \\
Acronym & 421 & Phonological deformation & 137 \\
Compounding & 373 & Blending / portmanteau & 124 \\
Part of speech -- adjective & 319 & Nominalization & 124 \\
Part of speech -- proper noun & 270 & Suffix -ιά & 101 \\
Part of speech -- verb & 234 & Diminutives & 100 \\
Figure of speech -- typological names & 212 & Figure of speech -- hyperbole & 95 \\
Figure of speech -- metaphor & 180 & Suffix -ίλα & 87 \\
\hline
\end{tabular}
\caption{Distribution of grammatical tags (top-20).}
\label{tab:grammar}
\end{table}

\subsubsection{Affective (Φόρτιση) mapping.}

These tags (see Table~\ref{tab:affective}) encode evaluative, emotional, and socially marked meanings, including offensiveness (e.g., \textit{insult}), bias (e.g., \textit{sexist}), and stance (e.g., \textit{irony}), often reflecting negative or taboo content. 
This confirms the central role of stance, humor, and social roles in slang. 
The distribution is skewed toward negative and transgressive language, with insults, sexist, and racist expressions dominating.

\begin{table}[ht]
\centering
\footnotesize
\begin{tabular}{lr@{\hspace{1.5em}}lr}
\hline
\textbf{Affective Category} & \textbf{Freq.} & \textbf{Affective Category} & \textbf{Freq.} \\
\hline
Insult & 2209 & Ethnic slur & 68 \\
Sexist & 1713 & Slogan-like & 56 \\
Racist & 605 & Command & 51 \\
Derogatory & 502 & Folk wisdom & 51 \\
Irony & 444 & Indignation & 50 \\
Disapproval & 203 & Admiration & 45 \\
Put-down / belittling & 177 & Disgust & 44 \\
Approval & 173 & Threat & 32 \\
Joking / humorous & 81 & Curse & 30 \\
Surreal & 76 & Snobbery & 26 \\
\hline
\end{tabular}
\caption{Distribution of affective tags  (top-20).}
\label{tab:affective}
\end{table}

\subsubsection{Foreign influence (Ξένες Επιρροές) mapping.}
These tags encode foreign origin or borrowing, including labels that specify the source language (e.g., \textit{English}) and more general tags for unspecified origin (e.g., \textit{foreign form}). As shown in Table~\ref{tab:foreign-influences}, most instances correspond to unspecified origin, followed by English as the dominant source. Contributions from Italian, Turkish, and French reflect the contact-driven nature of Greek slang, shaped by regional interaction and historical influences such as the Ottoman era.

\begin{table}[ht]
\centering
\footnotesize
\begin{tabular}{lr@{\hspace{1.5em}}lr}
\hline
\textbf{Foreign Influence Category} & \textbf{Freq.} & \textbf{Foreign Influence Category} & \textbf{Freq.} \\
\hline
Unspecified & 2425 & Spanish & 34 \\
English & 736 & Arabic / Persian & 29 \\
Foreign form & 222 & Back-borrowing & 28 \\
Hellenization & 216 & Slavic languages & 27 \\
Loanword & 207 & German & 21 \\
Italian & 142 & Latin & 17 \\
Turkish & 139 & Arvanitika & 10 \\
French & 138 & Russian & 9 \\
Calque / Loan translation & 89 & Albanian & 7 \\
Romani & 41 & Hebrew / Jewish language & 4 \\
\hline
\end{tabular}
\caption{Distribution of foreign-influence tags.}
\label{tab:foreign-influences}
\end{table}

\subsubsection{Regional (Περιοχή) mapping.}

This category (see Table \ref{tab:region}) captures geographical and dialectal variation, including major regions (e.g., \textit{Crete}) and urban centers (e.g., \textit{Athens}), reflecting both localized dialectal variation and urban sociolinguistic dynamics. Although less frequent, regional annotations capture dialectal diversity and the geographic distribution of slang, supporting the study of localized variation and regional identity.

\begin{table}[ht]
\centering
\footnotesize
\begin{tabular}{lr@{\hspace{1.5em}}lr}
\hline
\textbf{Region} & \textbf{Freq.} & \textbf{Region} & \textbf{Freq.} \\
\hline
Crete & 192 & Thessaloniki & 43 \\
Macedonia & 130 & Ioannina & 35 \\
Peloponnese & 129 & Athens & 30 \\
Cyprus & 98 & Piraeus & 22 \\
Aegean & 89 & Thrace & 20 \\
Ionian Islands & 75 & Pontus & 19 \\
Thessaly & 55 & Patras & 14 \\
Epirus & 52 & Dodecanese & 9 \\
Central Greece & 46 & Asia Minor & 6 \\
\multicolumn{2}{l}{} & Arta & 3 \\
\hline
\end{tabular}
\caption{Distribution of regional tags.}
\label{tab:region}
\end{table}

\subsubsection{Temporal (Χρονολόγηση) mapping.}

Table~\ref{tab:temporal} tags encode specific periods (e.g., \textit{1980s}) and broader markers (e.g., \textit{pre-war}). Although sparse, they capture the diachronic evolution of slang and generational patterns. Most expressions cluster in the late 20th century (1970s--1990s), a period of intensified innovation. Pseudo-archaic forms further show that slang draws on stylized historical usage, often for humorous or expressive effect.

\begin{table}[ht]
\centering
\footnotesize
\begin{tabular}{lr@{\hspace{1.5em}}lr}
\hline
\textbf{Temporal Category} & \textbf{Freq.} & \textbf{Temporal Category} & \textbf{Freq.} \\
\hline
1980s & 89 & 1940s & 28 \\
1970s & 69 & 19th century & 16 \\
Pseudo-archaic & 60 & 1950s & 15 \\
2000s & 58 & 1930s & 13 \\
1960s & 55 & Unknown, likely earlier centuries & 7 \\
1990s & 48 & Early 20th century & 2 \\
2010s--present & 43 & Middle Ages & 1 \\
Pre-war & 41 &  &  \\
\hline
\end{tabular}
\caption{Distribution of temporal tags.}
\label{tab:temporal}
\end{table}

\section{Taxonomy of Greek Slang}
\label{sec:taxonomy}

A central objective of this work is to map Greek slang expressions to a structured semantic taxonomy, enabling the systematic analysis of their meaning and usage. 
The \texttt{slang.gr} tags can be characterized as a \emph{folksonomy}, a bottom-up, user-driven tagging system with limited structure. These tags, curated by the user community, vary significantly in granularity and type. Some refer to semantic denotation while others encode non-semantic information. As a result, the tag space conflates multiple dimensions, including referential meaning, discourse function, register, chronology, and sociolinguistic metadata. Despite this heterogeneity, the tags provide a rich signal that can be leveraged to infer semantic structure.

To capture this distinction, we map the \texttt{slang.gr} folksonomy to a structured semantic taxonomy inspired by the Oxford Dictionary of Modern Slang \cite{ayto2010oxford} and used as an analytical tool rather than a fully evaluated linguistic resource. In addition to the Oxford semantic categories (A--L), we introduce an additional category, M, to capture non-semantic metadata, not covered by the original taxonomy, leading to a two-layer design:
\begin{enumerate}
    \item a \textbf{semantic layer} (A--L), capturing what a slang item refers to in conceptual terms, and
    \item a \textbf{metadata layer} (M), capturing non-semantic linguistic and contextual information.
\end{enumerate}

In this process, tags are treated as proxies for meaning and aligned with the conceptual structure of the taxonomy. A single tag may receive multiple labels across the semantic and metadata layers. This is necessary for slang, where a single tag may simultaneously encode, for example, semantic content, discourse function, linguistic form, and contextual metadata. For example, \textit{characterization of person} is mapped both semantically to \emph{B. People and Society} and metadata-wise to M7 (\emph{Referent}). 

The mapping is initially performed using an LLM\footnote{We used the paid version of OpenAI’s GPT-5.2 through the ChatGPT interface, accessed in April 2026.}, which is provided with the normalized tag inventory and the corresponding semantic and metadata taxonomy, and is tasked to assign each tag to one or more categories of the taxonomy in batches of 10 tags. The LLM was also asked to provide a confidence score for each label, allowing uncertain assignments to be easily inspected.

The resulting mappings were manually curated by the authors against their associated senses. The initial LLM output contained 672 assignments, including 409 semantic and 263 metadata mappings. Eight tags were flagged by the LLM as requiring further review, while 13 initially remained unmapped. The curation process involved careful examination of how the tags aligned with their associated senses. Of the 649 unique tags, 417 retained their original mappings, while 232 were manually revised, adding 274 mappings and removing 110. Following curation, all 649 tags received at least one label. The final taxonomy contains 836 assignments, comprising 501 semantic and 335 metadata mappings. Of these, 481 tags received a single label and 168 received multiple labels: 152 received two labels, 13 received three, and three received four. Tags receiving the maximum of four mappings combine semantic content with pragmatic stance or editorial metadata, as in \textit{καμάκι}. Two annotators reviewed the candidate mappings. Across 852 candidate annotation decisions before adjudication, they agreed on 824 and disagreed on 28, corresponding to 96.71\% raw agreement. Cohen's $\kappa$ was 0.932. After adjudicating disagreements and removing duplicate or rejected candidate assignments, the process yielded 836 final mappings.

\subsection{Semantic Layer (A--L)}

\begin{table*}[ht]
\centering
\footnotesize
\begin{tabular}{lcl@{\hspace{1.2em}}lcl}
\hline
\textbf{Code} & \textbf{Domain} & \textbf{\#} & 
\textbf{Code} & \textbf{Domain} & \textbf{\#} \\
\hline
A & The body and its functions & 12 & G & Behaviour, attitudes, and emotions & 50 \\
B & People and society & 24 & H & Thought and communication & 18 \\
C & Animals and Plants & 2 & I & The arts, entertainment, and the media & 6 \\
D & Sustenance and intoxication & 5 & J & Time and tide & 9 \\
E & Articles and substances & 8 & K & Location and movement & 10 \\
F & Money, commerce, and employment & 5 & L & Abstract qualities and states & 30 \\
\hline
\end{tabular}
\caption{Top-level semantic domains (A--L) with number of subcategories.}
\label{tab:semantic}
\end{table*}

The semantic layer (Table~\ref{tab:semantic}) follows the Oxford Dictionary of Modern Slang, comprising 12 high-level domains and 179 subcategories (depth = 2). Targeted extensions are introduced to better capture \texttt{slang.gr}. Specifically, category C is extended to include plants, while additional subcategories account for previously underrepresented domains such as \emph{E7. Technology \& Computing}, \emph{E8. Science}, \emph{I6. Art (General / Fine Arts)}, \emph{J9. Youth / Young}, and \emph{G50. Mental / emotional state (general)}. 

\subsection{Metadata Layer (M)}

The metadata layer (M) (Table~\ref{tab:metadata}) captures non-denotational properties of slang expressions, encoding formal, sociolinguistic, and interactional dimensions. It consists of eight top-level dimensions (M1--M8) and 54 subcategories, organized hierarchically with variable depth up to three levels as in M6 Pragmatic stance. Designed in a data-driven manner to reflect the \texttt{slang.gr} annotation scheme, it organizes heterogeneous tags into orthogonal and extensible axes for analyzing variation in form, usage, and social meaning.

\begin{table*}[ht]
\centering
\footnotesize
\begin{tabular}{llcp{5.5cm}@{\hspace{0.8em}}llcp{4.5cm}}
\hline
\textbf{Code} & \textbf{Dimension} & \textbf{\#} & \textbf{Description} &
\textbf{Code} & \textbf{Dimension} & \textbf{\#} & \textbf{Description} \\
\hline
M1 & Linguistic Form & 9 & Formal properties of the expression, including word formation (e.g., compounding) and grammatical behavior. &
M5 & Region & 4 & Geographical or dialectal distribution. \\

M2 & Register & 10 & Social and situational usage context (e.g., internet, youth, written language). &
M6 & Pragmatic stance & 13 & Speaker attitude and discourse function (evaluative, emotional, behavioral, moral). \\

M3 & Origin & 4 & Source-related information (e.g., borrowing, language influence). &
M7 & Referent & 7 & Target entity type (e.g., person, place, state). \\

M4 & Chronology & 3 & Temporal information (e.g., historical period, recency). &
M8 & \makecell[l]{Editorial /\\ Annotation} & 4 & Annotation-level or non-linguistic tags. \\
\hline
\end{tabular}
\caption{Metadata dimensions of the taxonomy (M1--M8) with number of leaf subcategories.}
\label{tab:metadata}
\end{table*}

\subsection{Discussion}

The taxonomy shows that Greek slang is inherently multidimensional, combining semantic content with sociolinguistic and pragmatic signals. By organizing the \texttt{slang.gr} folksonomy into a structured, multi-layer, multi-label representation, it enables the systematic analysis of meaning as well as form (M1), usage (M2), origin (M3), time (M4), region (M5), stance (M6), referent (M7), and annotation (M8). This highlights slang as a dynamic system shaped by community practices, identity, and expressive needs.

\begin{figure*}[t]
    \centering

    \begin{subfigure}[t]{0.32\textwidth}
        \vspace{0pt}
        \centering
        \includegraphics[width=\linewidth,height=3.2cm,keepaspectratio]{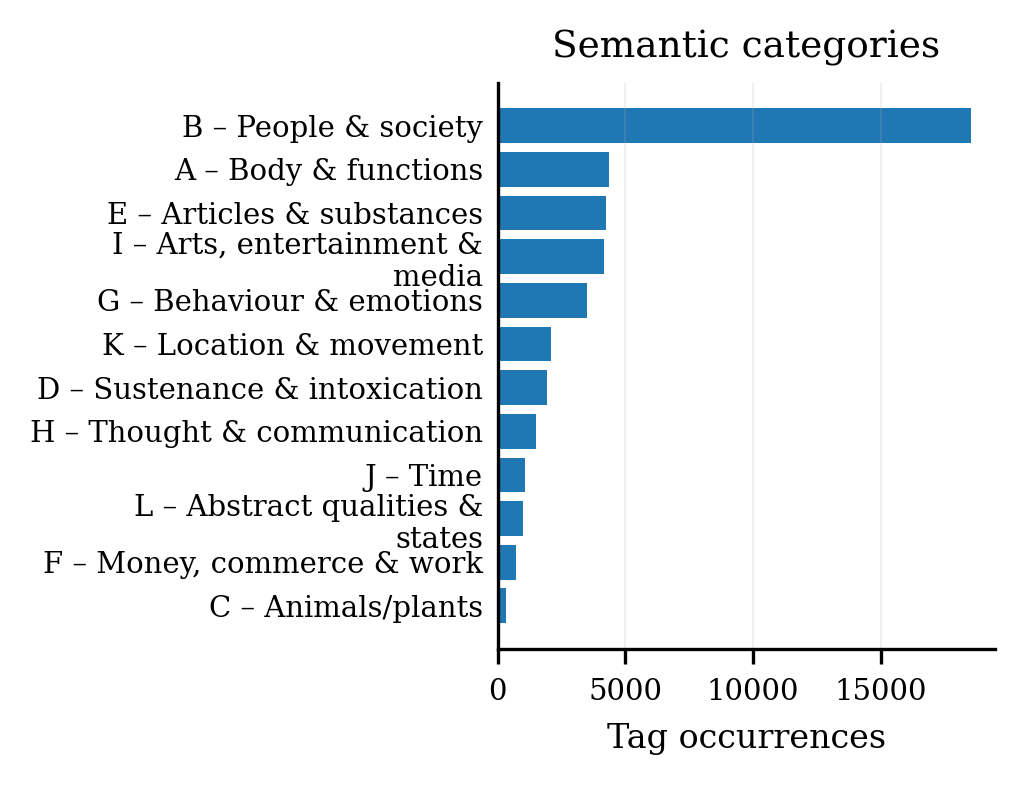}
        \caption{Overall semantic distribution}
        \label{fig:semantic_distribution}
    \end{subfigure}
    \hfill
    \begin{subfigure}[t]{0.32\textwidth}
        \vspace{0pt}
        \centering
        \includegraphics[width=\linewidth,height=3.2cm,keepaspectratio]{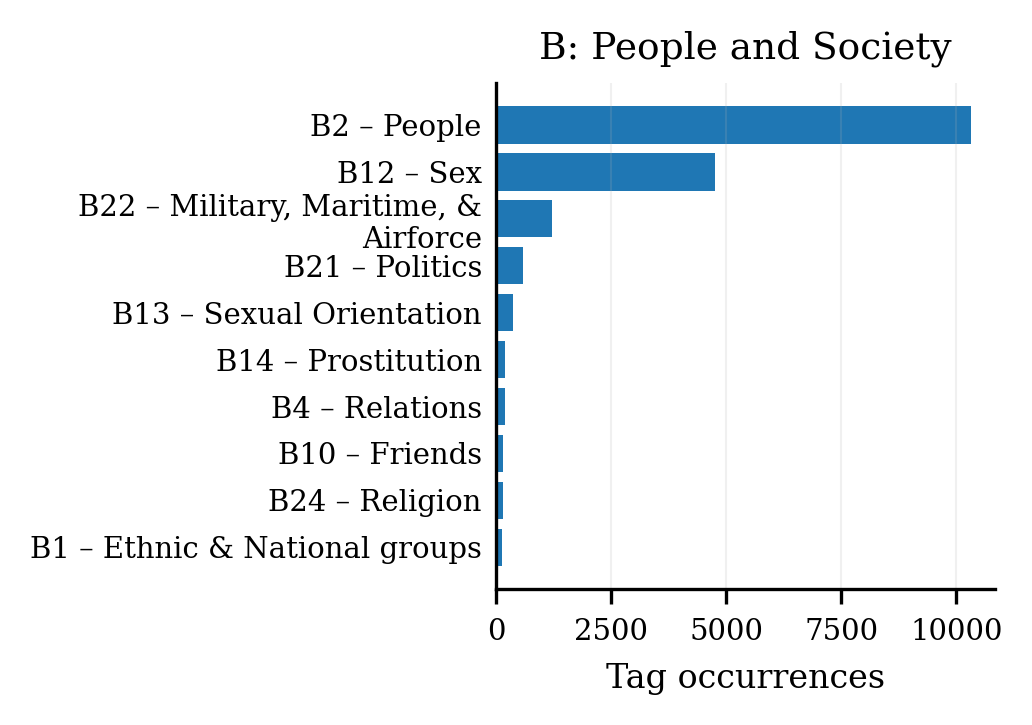}
        \caption{B: People and society}
        \label{fig:B_subcategories}
    \end{subfigure}
    \hfill
    \begin{subfigure}[t]{0.32\textwidth}
        \vspace{0pt}
        \centering
        \includegraphics[width=\linewidth,height=3.2cm,keepaspectratio]{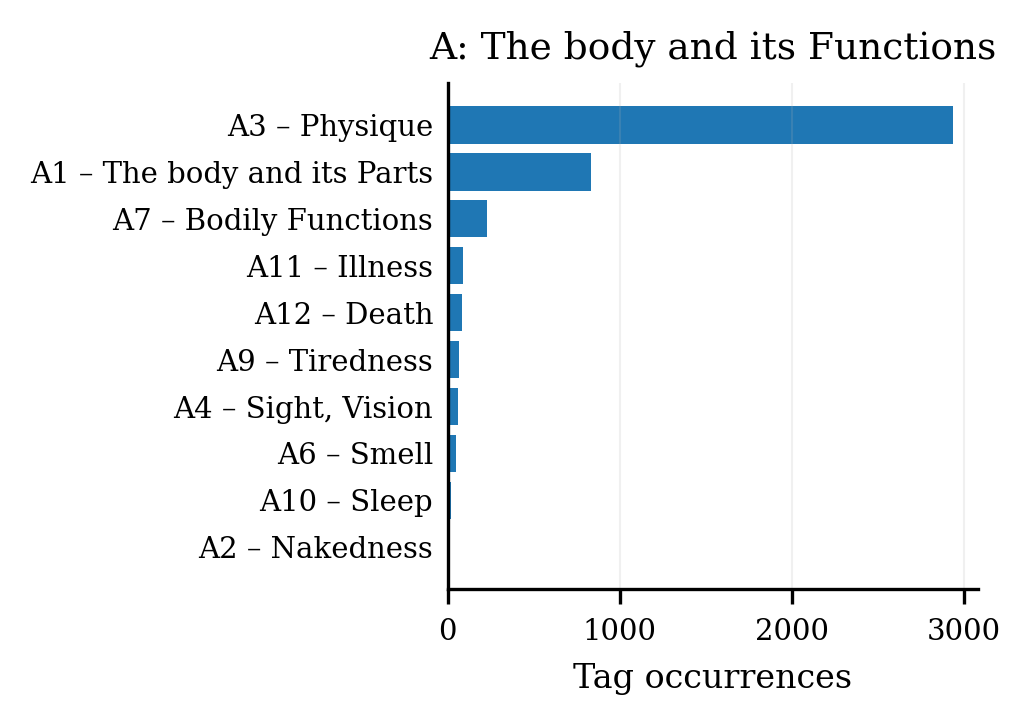}
        \caption{A: The body and its functions}
        \label{fig:A_subcategories}
    \end{subfigure}

    \vspace{0.5em}

    \begin{subfigure}[t]{0.32\textwidth}
        \vspace{0pt}
        \centering
        \includegraphics[width=\linewidth,height=3.2cm,keepaspectratio]{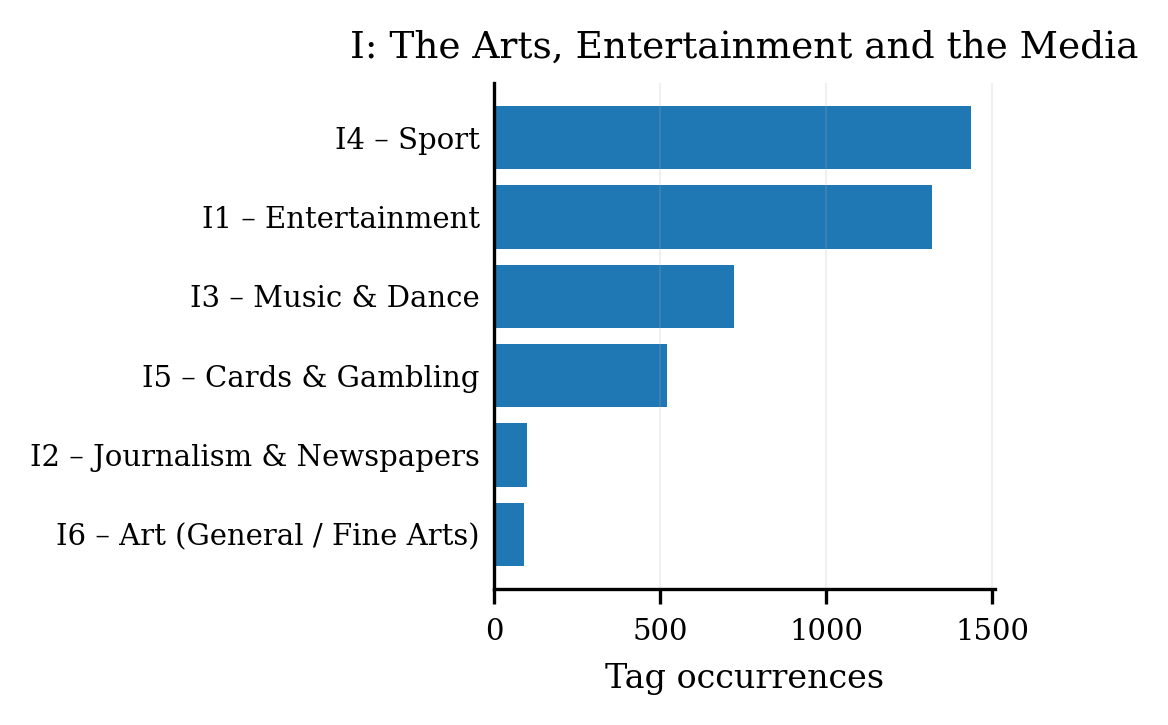}
        \caption{I: Arts, entertainment, and media}
        \label{fig:I_subcategories}
    \end{subfigure}
    \hfill
    \begin{subfigure}[t]{0.32\textwidth}
        \vspace{0pt}
        \centering
        \includegraphics[width=\linewidth,height=3.2cm,keepaspectratio]{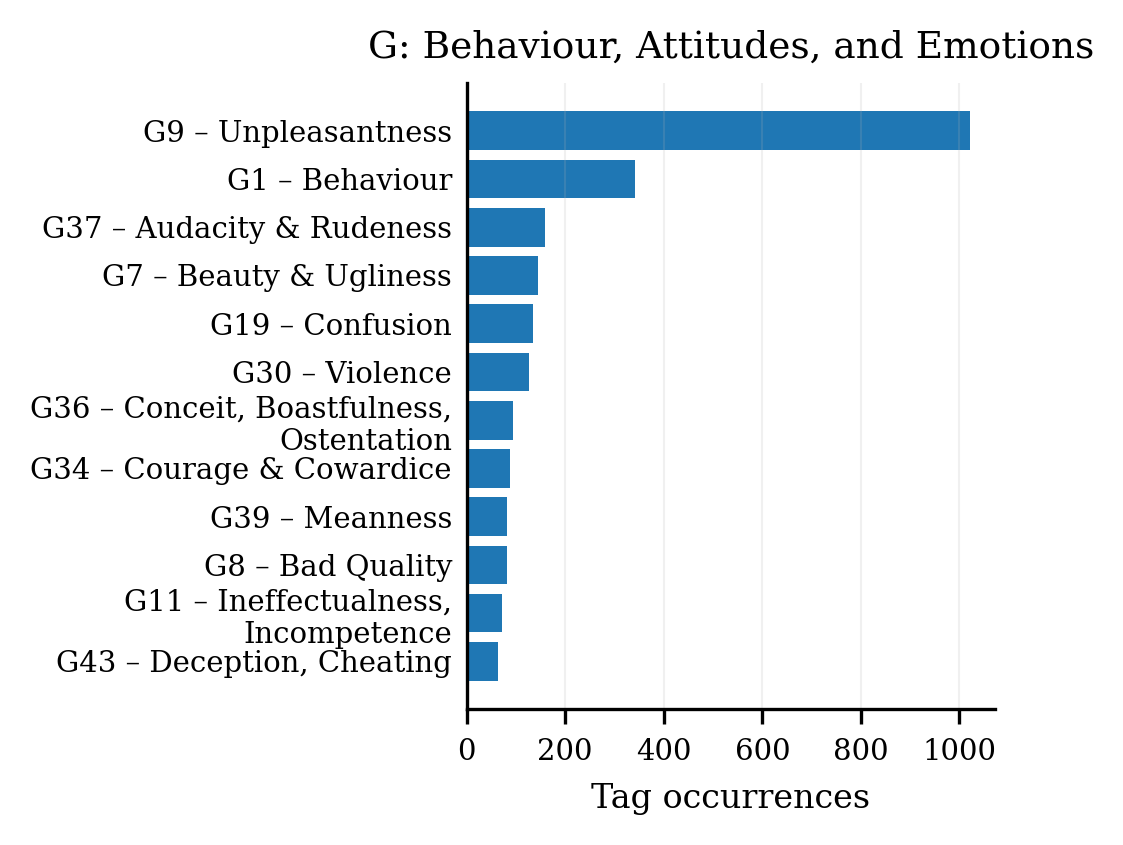}
        \caption{G: Behaviour, attitudes, and emotions}
        \label{fig:G_subcategories}
    \end{subfigure}
    \hfill
    \begin{subfigure}[t]{0.32\textwidth}
        \vspace{0pt}
        \centering
        \includegraphics[width=\linewidth,height=3.2cm,keepaspectratio]{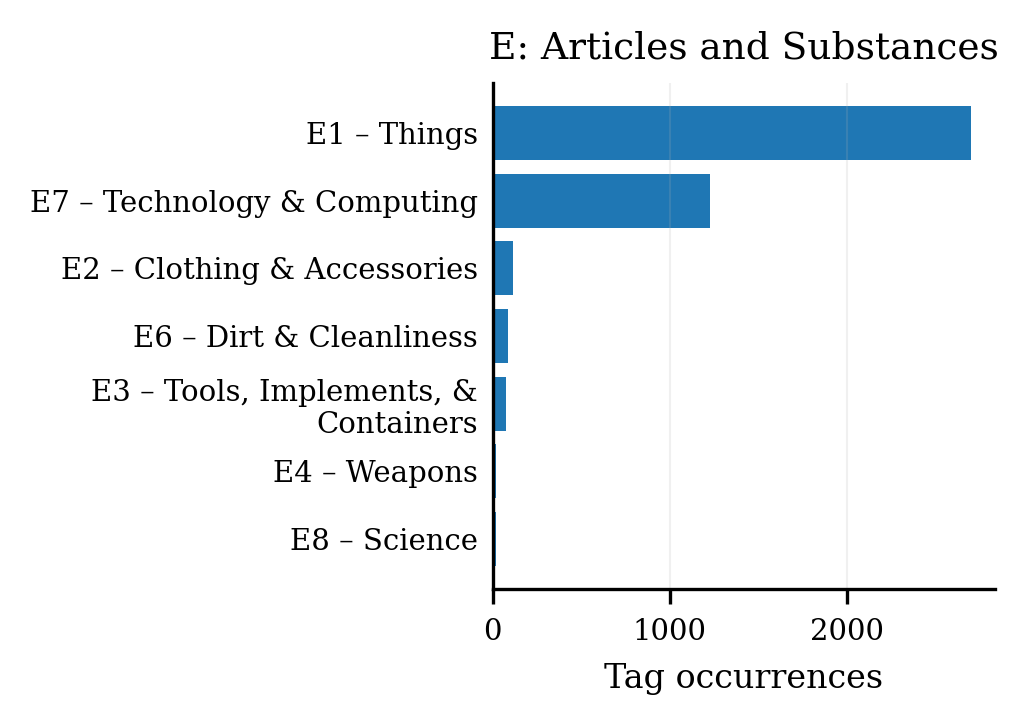}
        \caption{E: Articles and substances}
        \label{fig:E_subcategories}
    \end{subfigure}

    \caption{
    Overall distribution of semantic categories, followed by subcategory breakdowns for the most frequent domains.
    }
    \label{fig:combined_semantic}
\end{figure*}

Figure~\ref{fig:combined_semantic}(a) shows the distribution of lemma occurrences across the 12 Oxford semantic categories. The largest are \emph{B. People and society}, \emph{A. The body and its functions}, and \emph{I. Arts, entertainment, and the media}, followed by \emph{E. Articles and substances} and \emph{G. Behaviour, attitudes, and emotions}, indicating a focus on social actors, embodiment, culture, and evaluation.
Within \emph{B} (Fig.~\ref{fig:combined_semantic}(b)), subcategories such as \emph{B2. People} and \emph{B12. Sex} dominate, reflecting strong emphasis on person description and social characterization, with a highly skewed, long-tail distribution. Similarly, in \emph{A} (Fig.~\ref{fig:combined_semantic}(c)), \emph{A3. Physique}, \emph{A1. The body and its parts}, and \emph{A7. Bodily functions} dominate, highlighting embodiment and taboo-related domains.
In \emph{I} (Fig.~\ref{fig:combined_semantic}(d)), categories such as \emph{I4. Sports}, \emph{I1. Entertainment}, \emph{I3. Music and dance}, and \emph{I5. Cards and gambling} are prominent, with a more distributed structure. \emph{G} (Fig.~\ref{fig:combined_semantic}(e)) is similarly spread across subcategories such as \emph{G9. Unpleasantness}, \emph{G1. Behaviour}, \emph{G37. Audacity and rudeness}, and \emph{G7. Beauty and ugliness}. The distribution is less sharply peaked, with multiple subcategories contributing substantially, indicating broader evaluative variation. Finally, \emph{E} (Fig.~\ref{fig:combined_semantic}(f)) highlights the role of material and technological domains.

\captionsetup[subfigure]{labelformat=parens}

\captionsetup[subfigure]{labelformat=parens}

\begin{figure*}[t]
    \centering

    \begin{subfigure}[t]{0.32\textwidth}
        \vspace{0pt}
        \centering
        \includegraphics[width=\linewidth,height=3.6cm,keepaspectratio]{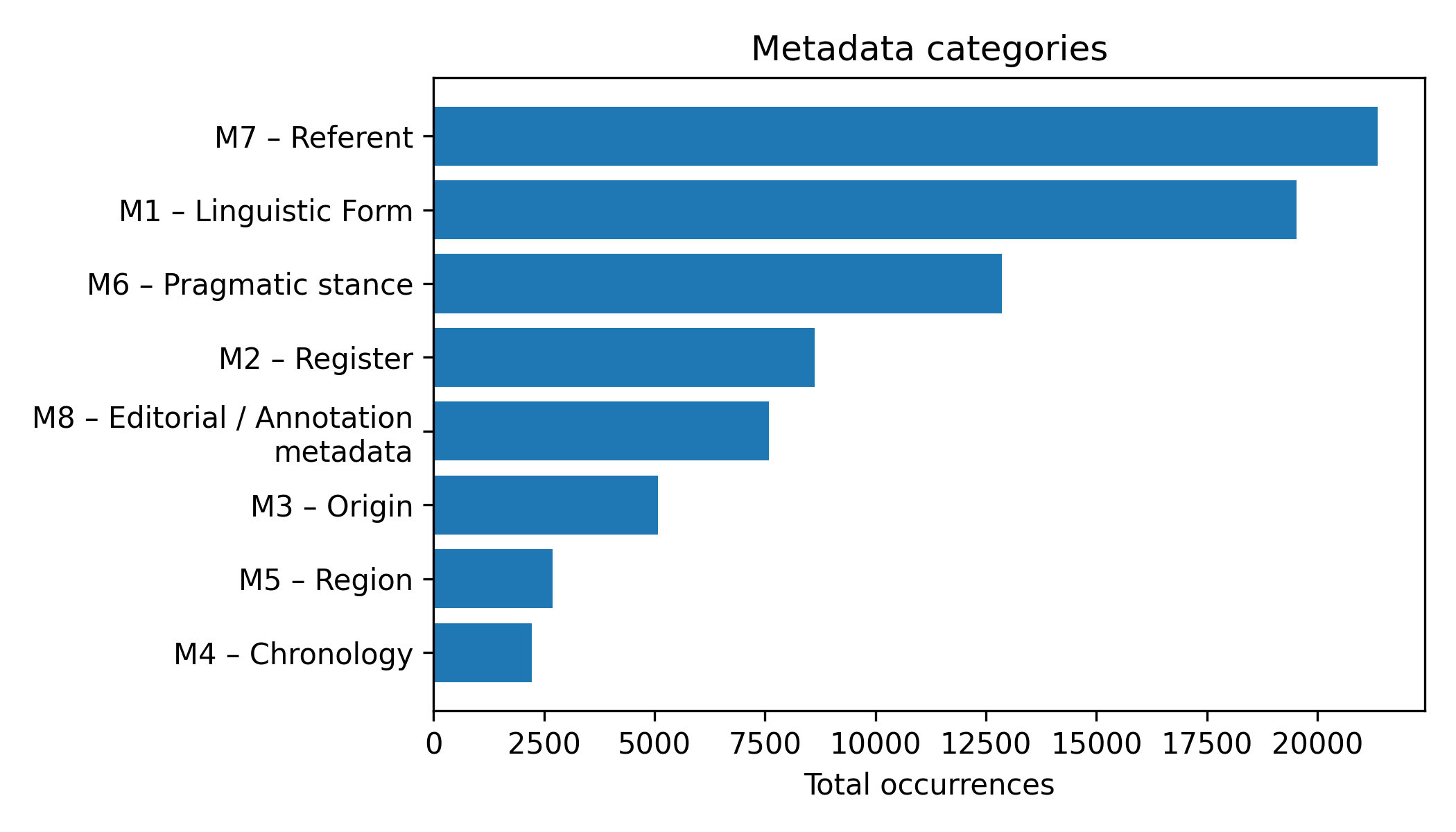}
        \caption{Overall metadata distribution}
        \label{fig:metadata_distribution}
    \end{subfigure}
    \hfill
    \begin{subfigure}[t]{0.32\textwidth}
        \vspace{0pt}
        \centering
        \includegraphics[width=\linewidth,height=3.6cm,keepaspectratio]{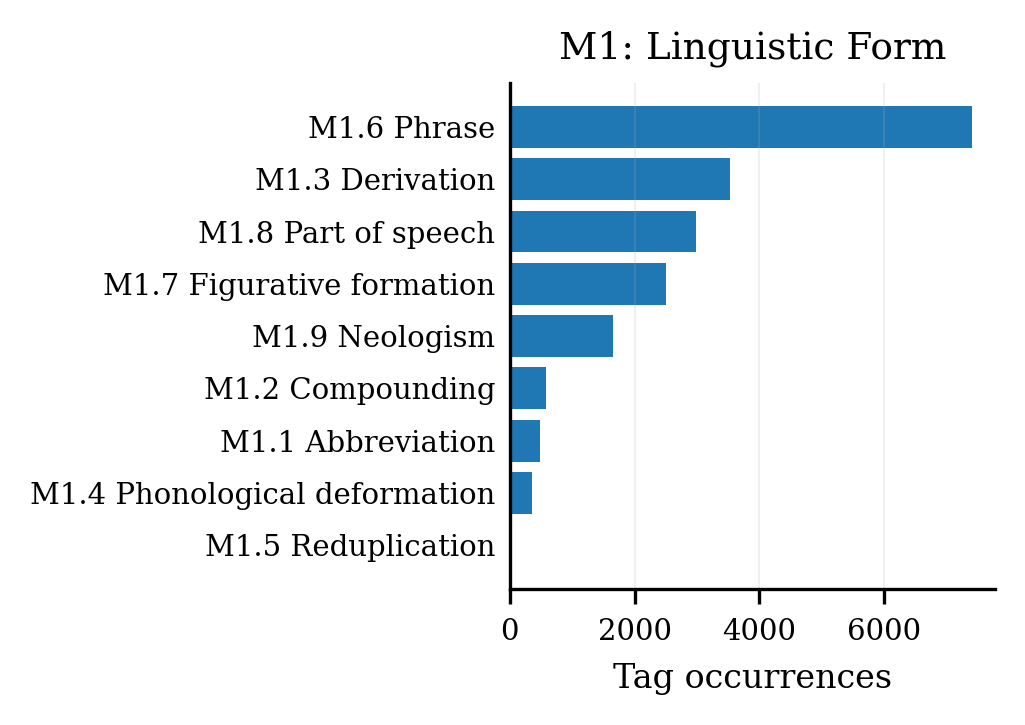}
        \caption{M1: Linguistic Form}
        \label{fig:M1_metadata}
    \end{subfigure}
    \hfill
    \begin{subfigure}[t]{0.32\textwidth}
        \vspace{0pt}
        \centering
        \includegraphics[width=\linewidth,height=3.6cm,keepaspectratio]{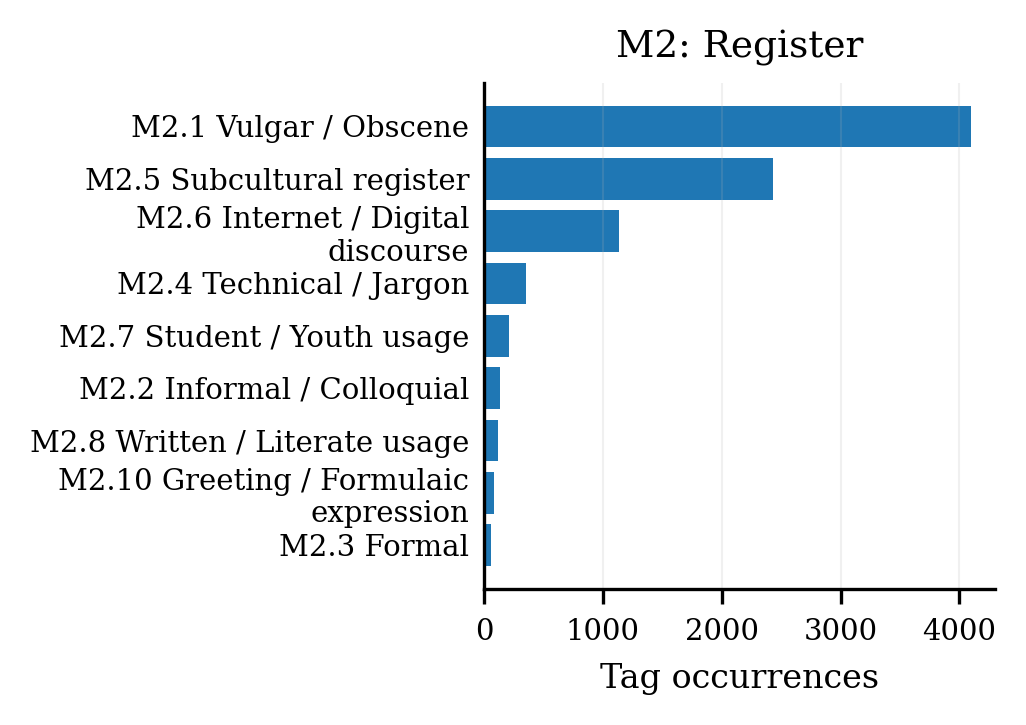}
        \caption{M2: Register}
        \label{fig:M2_metadata}
    \end{subfigure}

    \vspace{0.5em}

    \begin{subfigure}[t]{0.32\textwidth}
        \vspace{0pt}
        \centering
        \includegraphics[width=\linewidth,height=3.6cm,keepaspectratio]{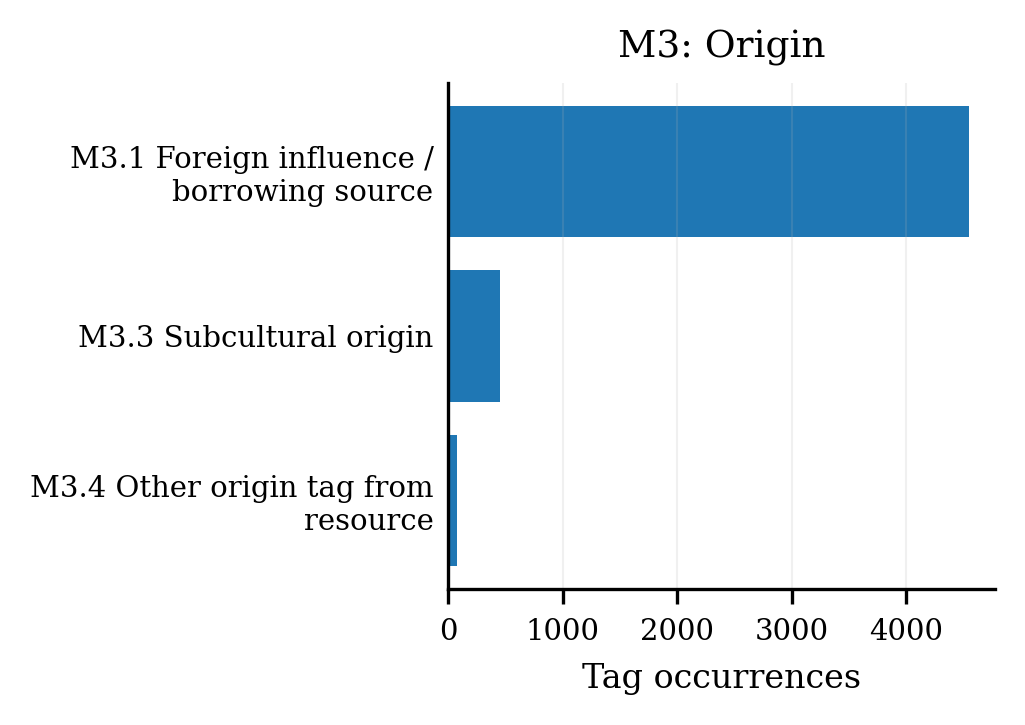}
        \caption{M3: Origin}
        \label{fig:M3_metadata}
    \end{subfigure}
    \hfill
    \begin{subfigure}[t]{0.32\textwidth}
        \vspace{0pt}
        \centering
        \includegraphics[width=\linewidth,height=3.6cm,keepaspectratio]{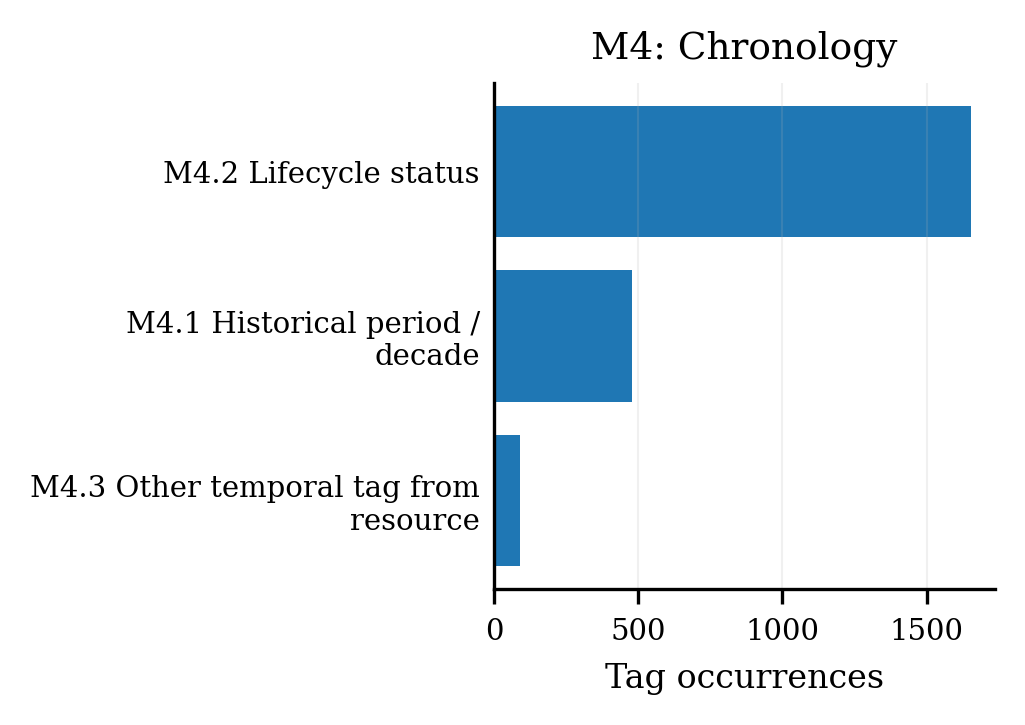}
        \caption{M4: Chronology}
        \label{fig:M4_metadata}
    \end{subfigure}
    \hfill
    \begin{subfigure}[t]{0.32\textwidth}
        \vspace{0pt}
        \centering
        \includegraphics[width=\linewidth,height=3.6cm,keepaspectratio]{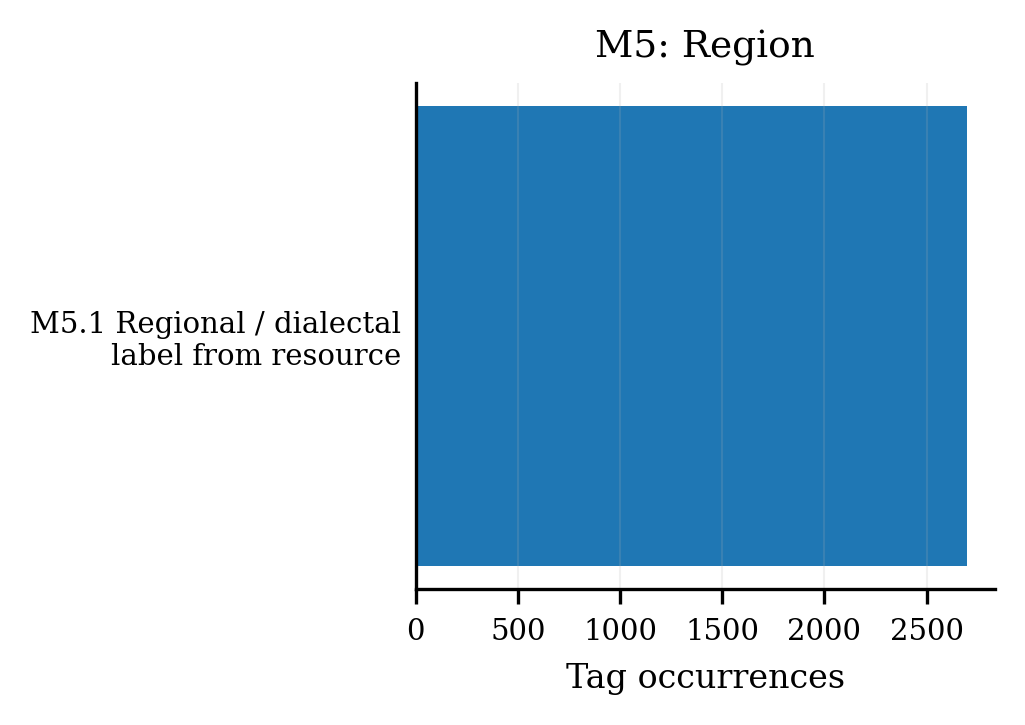}
        \caption{M5: Region}
        \label{fig:M5_metadata}
    \end{subfigure}

    \vspace{0.5em}

    \begin{subfigure}[t]{0.32\textwidth}
        \vspace{0pt}
        \centering
        \includegraphics[width=\linewidth,height=3.6cm,keepaspectratio]{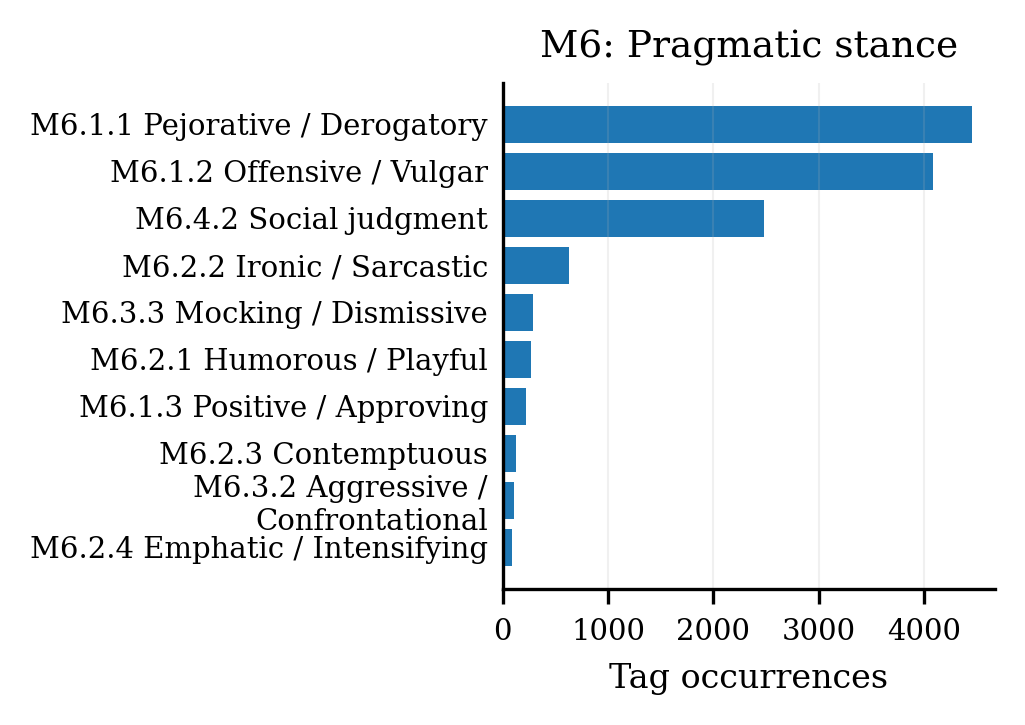}
        \caption{M6: Pragmatic Stance}
        \label{fig:M6_metadata}
    \end{subfigure}
    \hfill
    \begin{subfigure}[t]{0.32\textwidth}
        \vspace{0pt}
        \centering
        \includegraphics[width=\linewidth,height=3.6cm,keepaspectratio]{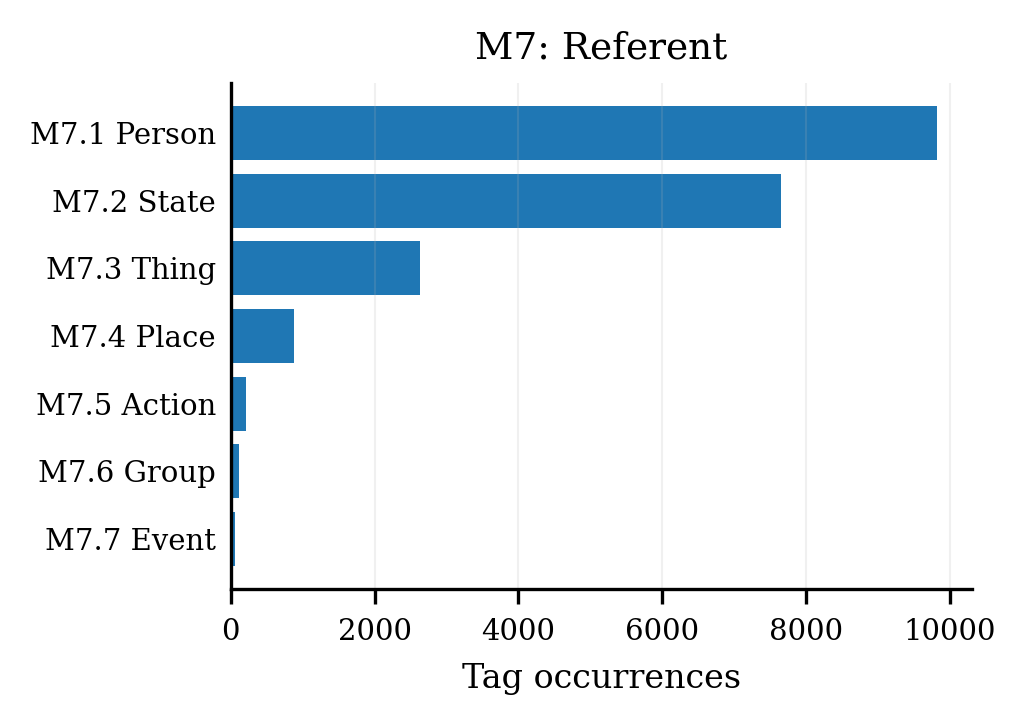}
        \caption{M7: Referent}
        \label{fig:M7_metadata}
    \end{subfigure}
    \hfill
    \begin{subfigure}[t]{0.32\textwidth}
        \vspace{0pt}
        \centering
        \includegraphics[width=\linewidth,height=3.6cm,keepaspectratio]{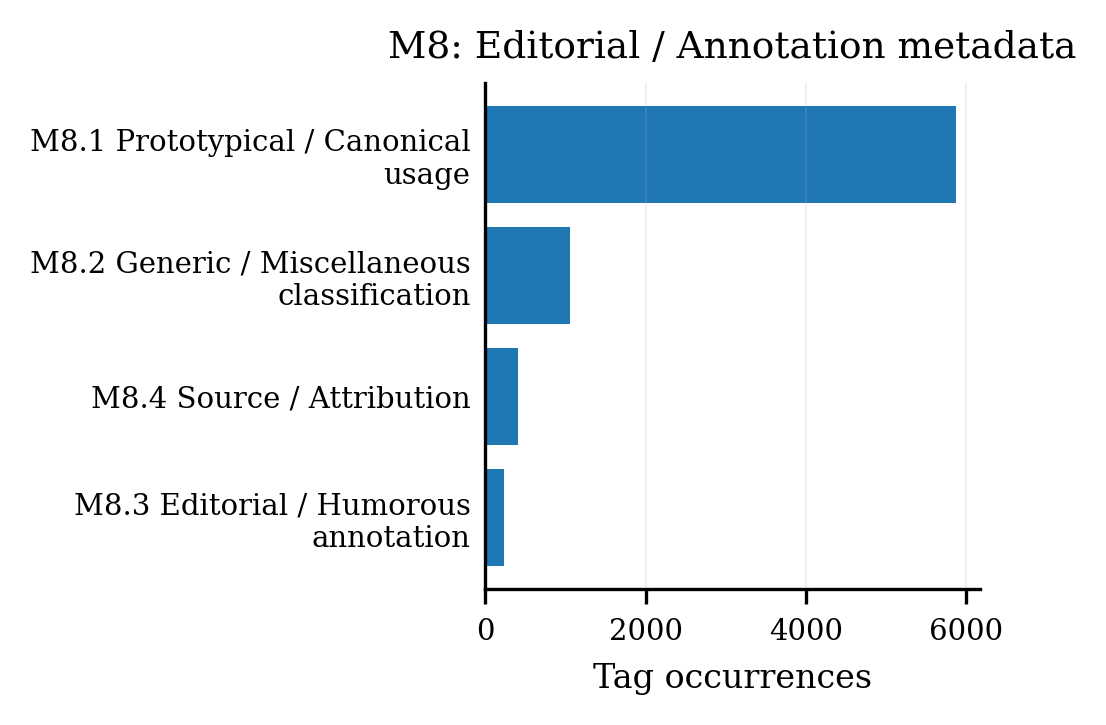}
        \caption{M8: Editorial / Annotation}
        \label{fig:M8_metadata}
    \end{subfigure}

    \vspace{0.5em}

    \caption{
    Overall distribution of metadata dimensions, followed by subcategory breakdowns.
    }
    \label{fig:metadata_subcategories_heatmap}
\end{figure*}

The metadata layer reveals the structure of sociolinguistic and pragmatic variation. As shown in Fig.~\ref{fig:metadata_subcategories_heatmap}(a), referent (M7), linguistic form (M1), pragmatic stance (M6), and register (M2) are the most prominent. The subcategory breakdowns (Fig.~\ref{fig:metadata_subcategories_heatmap}(b--i)) show that M1 emphasizes \emph{multiword expressions} and \emph{morphological processes}, while M2 reflects strong \emph{register differentiation}, including \emph{vulgar}, \emph{subcultural}, and \emph{internet-mediated} usage. M3 captures language contact and borrowing, while M4 (temporal) and M5 (regional) are sparser but provide contextual grounding. M6 exhibits rich internal structure, with \emph{pejorative/derogatory}, \emph{offensive/vulgar}, \emph{social judgment}, and \emph{ironic/sarcastic} uses being most populated, highlighting evaluation, offensiveness, and social positioning as central features of slang. M7 reinforces  \emph{person-centered expressions}, aligning with category B, while M8 captures annotation-level tags.

\begin{figure}[ht]
    \centering
    \includegraphics[width=0.65\linewidth]{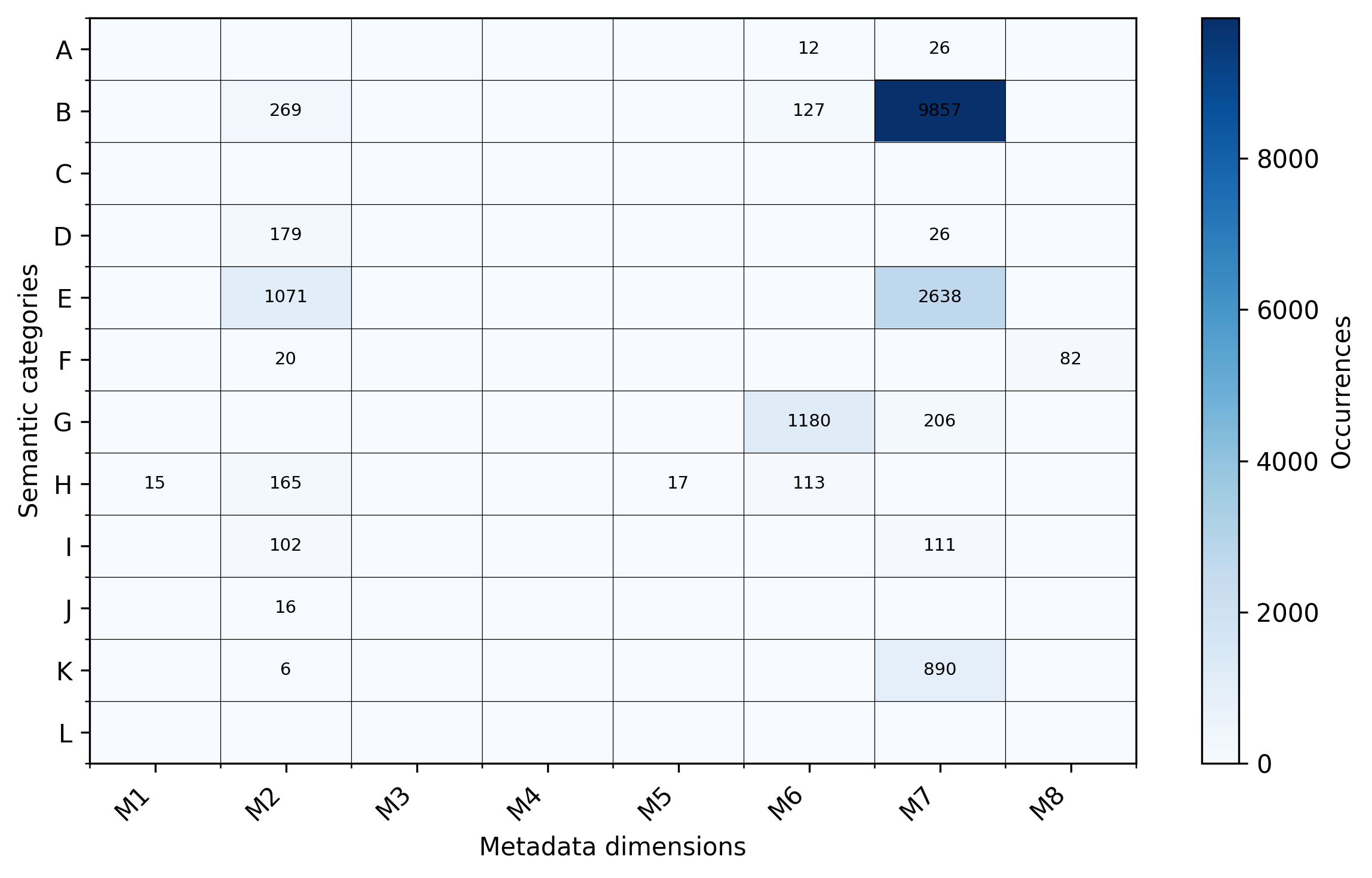}
    \caption{Semantic--Metadata heatmap}
    \label{fig:semantic_metadata_heatmap}
\end{figure}

The semantic--metadata co-occurrence heatmap (Fig.~\ref{fig:semantic_metadata_heatmap}) highlights a few strong associations between semantic domains and metadata dimensions. In particular, \emph{B. People and society} aligns primarily with \emph{M7 (Referent)} and secondarily with \emph{M6 (Pragmatic stance)}, reflecting its focus on person-centered and evaluative language. \emph{E. Articles and substances} and \emph{K. Location and movement} are also strongly associated with \emph{M7}, indicating their referential role in denoting concrete and spatial entities.  In contrast, \emph{G. Behaviour, attitudes, and emotions} is most strongly linked to \emph{M6}, showing the central role of evaluation and affect. Other associations are weaker, showing semantics alone is insufficient, as slang emerges from the interaction of denotation, social context, and expressive function.

\section{\texttt{slang.gr} Community Analysis}
\label{sec:community}

\texttt{slang.gr} records contributor activity over time, enabling us to study it as both a crowdsourced dictionary and an online community. We analyze participation, retention, and temporal dynamics, then construct tag-based, semantic, and metadata-driven user similarity graphs to examine community structure, and finally examine comment-based interactions, incorporating sentiment signals to define a community-based measure of definition quality.

\subsection{Users}

The platform includes three user types: \textit{creators} (write definitions), \textit{commenters} (only comment), and \textit{hybrids} (both). Of the 17,942 registered users, 3,358 (18.7\%) are active: 1,687 are creators (50.2\%), 642 commenters (19.1\%), and 1,029 hybrids (30.6\%). Across 101,957 comments (avg. 26.4 tokens, 61.0 comments per commenting user), the three most prolific users alone account for over one-fifth of the total. Demographic data is sparse. Gender is mostly unreported (92.4\% overall, 82.5\% active), with more male (5.4\% overall, 14.0\% active) than female users (2.2\% overall, 3.5\% active). Age disclosure is 35.2\% overall and 51.4\% for active users, with median ages of 39 and 34 years, respectively. 
Administrators may ban users for violating platform rules. As of April 2026, 99 users (0.6\%) were banned, of whom 30 were active (1 creator, 6 commenters, and 23 hybrids). Among them, 24 contributed 3,382 definitions and 29 posted 20,787 comments, while 69 never contributed, suggesting that many bans occurred before observable contribution activity, although removed activity may not be represented in the dataset. Table~\ref{tab:top-contributors} presents the top contributors by definition count, revealing a highly skewed distribution. The most active user has authored 2,554 definitions (more than the next two combined) and 7,236 comments. This is reflected in high Gini coefficients (0.812 for definitions and 0.952 for comments), exceeding typical crowdsourced platforms (e.g., a Gini ≈ 0.68 on X's Community Notes~\cite{razuvayevskaya2025gini}) and comparable to Wikipedia's edit inequality (Gini ≈ 0.95 for edits per user~\cite{pilati2025gini}). Notably, 4 of the top 10 creators are banned.

\begin{table}[htbp]
\centering
\footnotesize
\caption{Top 10 contributors by definition count}
\label{tab:top-contributors}
\begin{tabular}{llrr @{\hspace{1.5em}} llrr}
\toprule
\multicolumn{4}{c}{\textbf{Top 1--5}} & \multicolumn{4}{c}{\textbf{Top 6--10}} \\
\cmidrule(r){1-4} \cmidrule(l){5-8}
\textbf{User} & \textbf{Role} & \textbf{Def.} & \textbf{Com.} &
\textbf{User} & \textbf{Role} & \textbf{Def.} & \textbf{Com.} \\
\midrule
7659 & Hyb. & 2,554 & 7,236 & 17701 & Hyb. (banned) & 559 & 2,162 \\
6495 & Hyb. & 1,257 & 6,740 & 816 & Hyb. & 455 & 2,673 \\
3564 & Hyb. (banned) & 798 & 243 & 5942 & Hyb. & 386 & 3,052 \\
15610 & Hyb. (banned) & 682 & 8,311 & 3013 & Hyb. & 368 & 501 \\
5102 & Hyb. (banned) & 607 & 6,400 & 11755 & Hyb. & 348 & 3,005 \\
\bottomrule
\end{tabular}
\end{table}

User retention (time between first and last activity) could be computed for 3,330 contributors and shows a highly skewed engagement pattern, with 42.9\% contributing only once. For returning users, the median lifespan is 10 days, with 34.2\% completing all activity within a single day, while only 463 users (13.9\%) remained active for over a year.
Collaboration (multiple users per lemma) is rare with 90.5\% of lemmas being solo-defined, and the rest average 2.39 users.
This is reinforced by low engagement per lemma. Even the most active term (τάπα) received 20 definitions and comments.

\begin{table}[t!]
\centering
\footnotesize
\caption{Activity at launch, peak and year intervals.}
\label{tab:temporal-user}
\begin{tabular}{lrrrrrrr}
\toprule
 & \multicolumn{1}{c}{2006} 
 & \multicolumn{1}{c}{2007--2008} 
 & \multicolumn{1}{c}{2009} 
 & \multicolumn{1}{c}{2010--2014} 
 & \multicolumn{1}{c}{2015--2019} 
 & \multicolumn{1}{c}{2020--2026}  \\
\midrule
Definitions & 710 & 7,181 & 6,949 & 9,985 & 2,468 & 1,091  \\
Comments & -- & 8,645 & 38,121 & 46,247 & 8,680 & 264  \\
\bottomrule
\end{tabular}
\end{table}

\subsection{Temporal Analysis}
The platform launched in 2006 and peaked in 2009 with 6,949 definitions and 38,121 comments (Table~\ref{tab:temporal-user}). Activity declined sharply thereafter. Definitions fell to 53 in 2020 before a modest revival in 2024–2025 (346 and 303), while comments collapsed from 38,121 to under 50 annually after 2020 and never recovered. This trajectory aligns with the online community lifecycle model reported in~\cite{iriberri2009lifecycle}.

\subsection{User Communities (UC) Analysis}

\subsubsection{Tag Similarity UC}

We construct a user similarity graph by representing each user as a tag vector over their definitions. Let $|c_{u,i}|$ denote the frequency of tag $i$ for user $u$. Counts are TF--IDF weighted to reduce the influence of common tags and emphasize distinctive ones. Vectors are $\ell_2$-normalized and compared using cosine similarity. To ensure sparsity, we build a $k$-nearest neighbor graph ($k=20$) with a minimum similarity threshold of $0.05$. Users with fewer than three tagged definitions and tags appearing in less than two users were excluded to reduce noise. Edges are undirected and weighted by cosine similarity. The final graph contains only 1,064 users out of the 2,716 creator and hybrid users and 18,012 edges, forming a single sparse connected component.

\begin{figure}[t]
    \centering
    \begin{minipage}[t]{0.48\columnwidth}
        \centering
        \includegraphics[width=\linewidth]{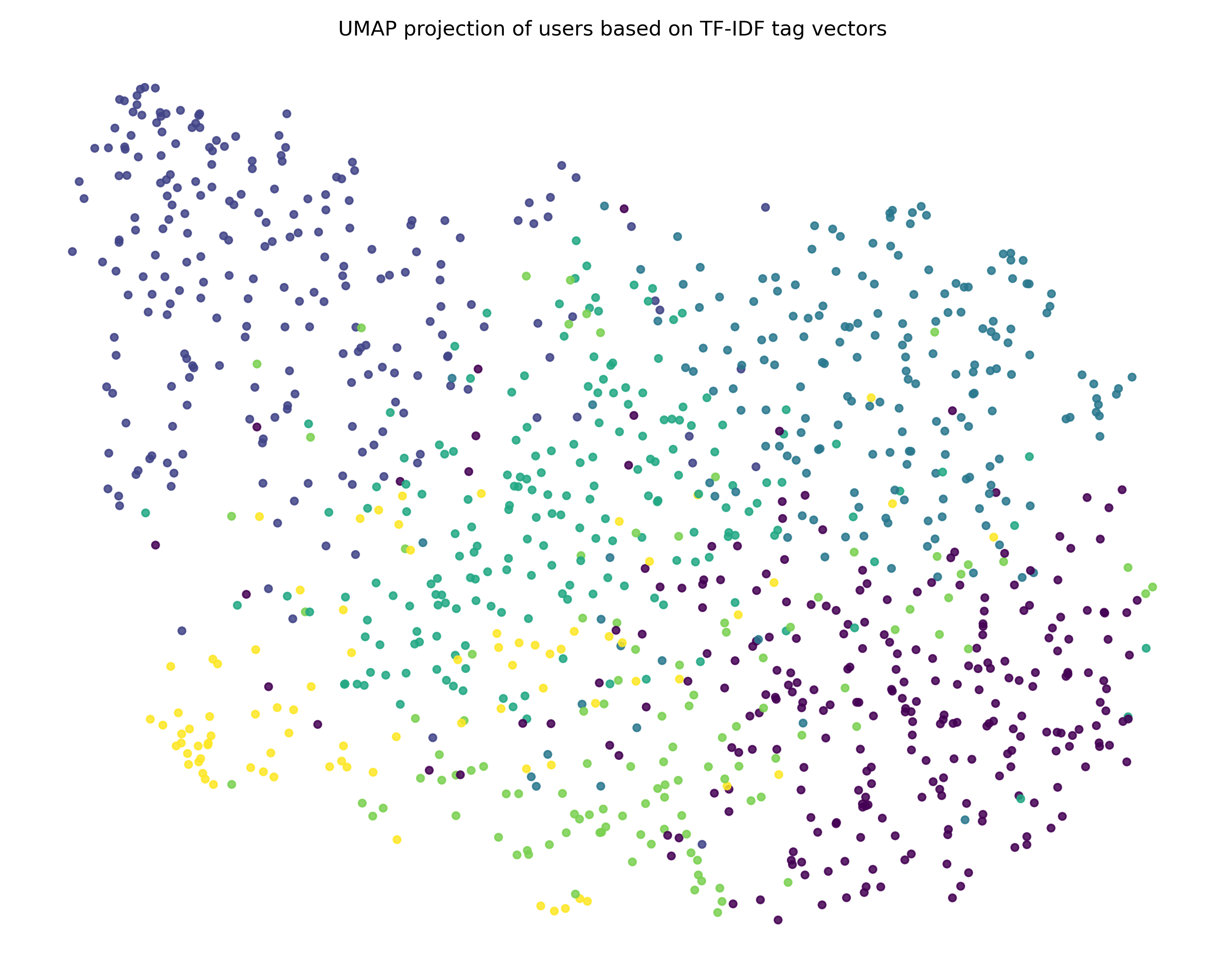}
        \caption*{(a) UMAP projection}
    \end{minipage}
    \hfill
    \begin{minipage}[t]{0.48\columnwidth}
        \centering
        \includegraphics[width=\linewidth]{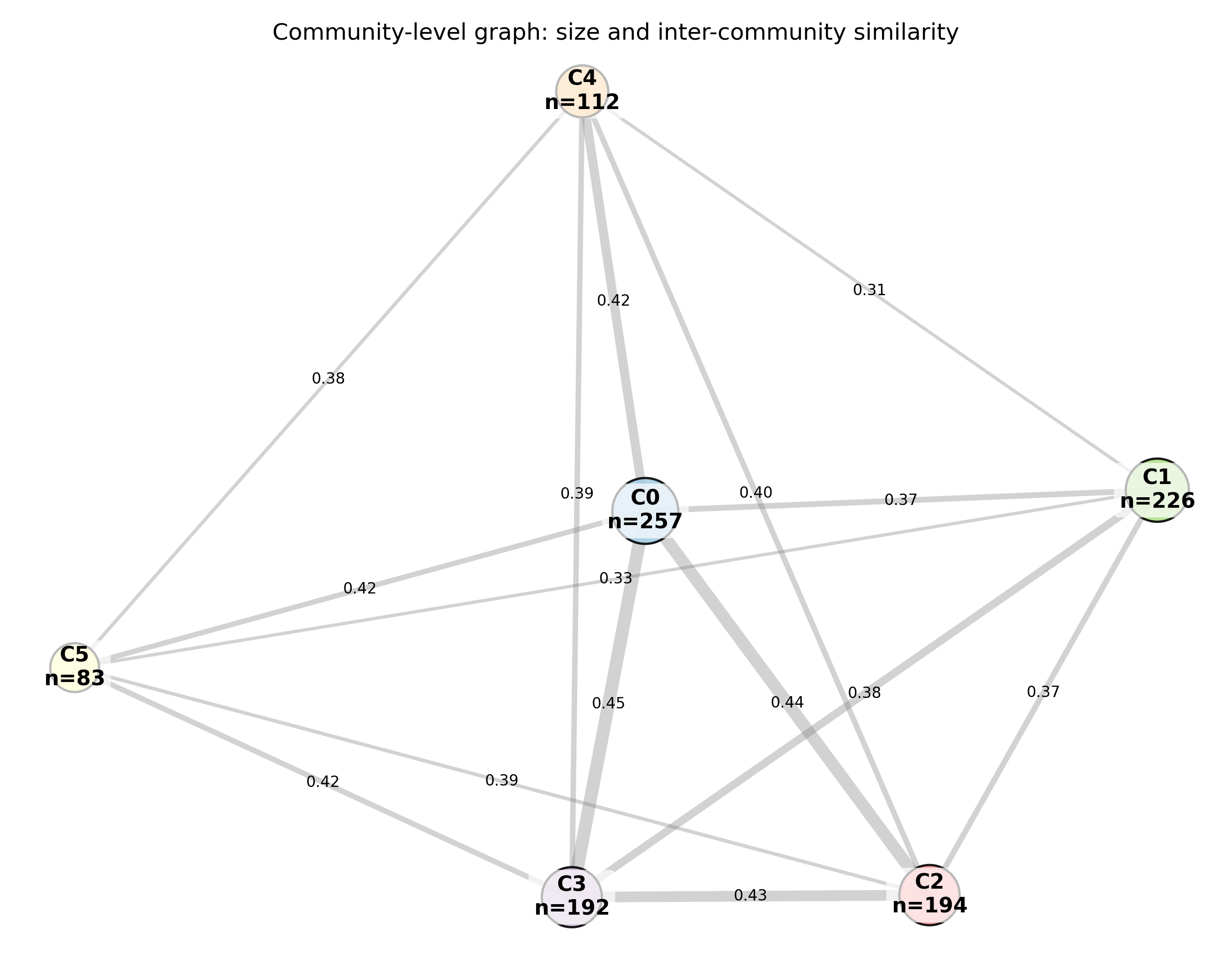}
        \caption*{(b) Community-level graph}
    \end{minipage}
    \caption{User communities based on tag similarity.}
    \label{fig:tag_umap_community}
\end{figure}

\begin{figure}[t]
    \centering
    \includegraphics[width=\columnwidth]{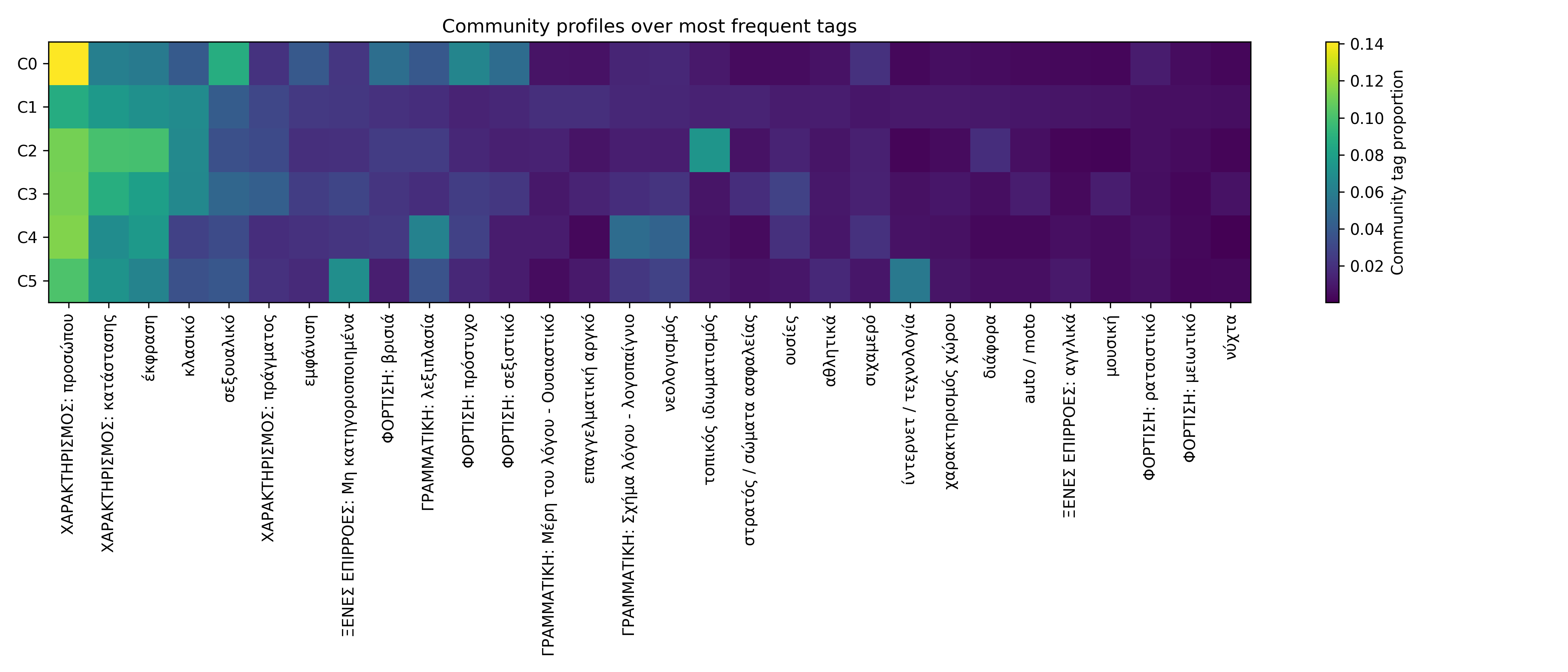}
    \caption{Tag-based community distributions over tags.}
    \label{fig:community_heatmap}
\end{figure}

We deployed the Leiden algorithm~\cite{traag2019leiden} (resolution  $1.0$) for community detection, yielding  six communities with weighted modularity $0.37$, indicating moderate community separation, similar to other social networks (e.g., 0.419 for the karate club network ~\cite{newman2006modularity}). Figure~\ref{fig:tag_umap_community}(a) shows a two-dimensional UMAP projection of users, revealing both compact clusters and transitional regions with mixed tagging behavior.  Fig.~\ref{fig:tag_umap_community}(b) presents the community-level graph, where the node diameter represents the community size and edge thickness reflects the inter-community similarity. Communities are strongly interconnected, with C0, C2, C3, and C4 more similar to each other than to C1 and C5.
Fig.~\ref{fig:community_heatmap} shows the distributions of the original tags across communities. The community profiles are mainly driven by person and state characterization, and differ in evaluative language, word formation processes, and more localized focus on domains such as technology and regional usage. A coherence measure based on intra-community cosine similarity relative to a random baseline indicates non-random grouping (coherence lift between $1.1$ and $1.44$, mean $1.23$).

Beyond aggregate structure, we analyze user-level variation by defining roles using percentile-based thresholds over three metrics: strength, betweenness centrality, and tag entropy. Strength is the total weight of a user’s connections, betweenness centrality measures how often a user lies on shortest paths (using $1-\mathrm{similarity}$ as distance), and tag entropy is the Shannon entropy of their tag distribution.  The analysis focuses on users retained in the main filtered graph. Thresholds are computed from the empirical distributions of these metrics within each graph. Core users are the top 10\% in strength, peripheral users the bottom 25\%, bridge users the top 5\% in betweenness, specialists the bottom 25\% in entropy, and generalists the top 10\%. This percentile-based approach allows comparability across graphs. Roles are not mutually exclusive.

Table~\ref{tab:user_roles_comparison} summarizes role distributions and banned-user proportions. Peripheral users show low banned rates (0.8\%), while generalists have the highest (6.5\%). Core users are moderate (2.8\%), bridge users show none, and specialists remain low (0.4\%), indicating that broader participation aligns with moderation signals.

\begin{table*}[t]
\centering
\footnotesize
\begin{tabular}{l c | cc | cc | cc}
\hline
 & & \multicolumn{2}{c}{\textbf{Tags}} & \multicolumn{2}{c}{\textbf{Semantic}} & \multicolumn{2}{c}{\textbf{Metadata}} \\
\textbf{Role} & \textbf{Threshold} & \textbf{Kept (\%)} & \textbf{Banned (\%)} & \textbf{Kept (\%)} & \textbf{Banned (\%)} & \textbf{Kept (\%)} & \textbf{Banned (\%)} \\
\hline
Core & top 10\% strength
& 107 (10.1\%) & 3 (2.8\%)
& 74 (10.0\%) & 4 (5.4\%)
& 92 (10.1\%) & 2 (2.2\%) \\

Peripheral & bottom 25\% strength
& 266 (25.0\%) & 2 (0.8\%)
& 185 (25.0\%) & 4 (2.2\%)
& 229 (25.1\%) & 0 (0.0\%) \\

Specialists & bottom 25\% entropy
& 267 (25.1\%) & 1 (0.4\%)
& 185 (25.0\%) & 0 (0.0\%)
& 230 (25.2\%) & 1 (0.4\%) \\

Generalists & top 10\% entropy
& 107 (10.1\%) & 7 (6.5\%)
& 74 (10.0\%) & 8 (10.8\%)
& 92 (10.1\%) & 8 (8.7\%) \\

Bridge & top 5\% betweenness
& 54 (5.1\%) & 0 (0.0\%)
& 37 (5.0\%) & 1 (2.7\%)
& 46 (5.0\%) & 1 (2.2\%) \\
\hline
\textbf{Retained users} & --
& 1064 (39.2\%) & 21 (2.0\%)
& 740 (27.3\%) & 17 (2.3\%)
& 914 (33.7\%) & 21 (2.3\%) \\
\hline
\end{tabular}
\caption{User roles across graphs. \textit{Kept} (\%) per retained graph, \textit{Banned} (\%) per role, retained users per total (2,716)}
\label{tab:user_roles_comparison}
\end{table*}

\subsubsection{Semantic Taxonomy Similarity UC}

\begin{figure}[t]
    \centering
    \begin{minipage}[t]{0.48\columnwidth}
        \centering
        \includegraphics[width=\linewidth]{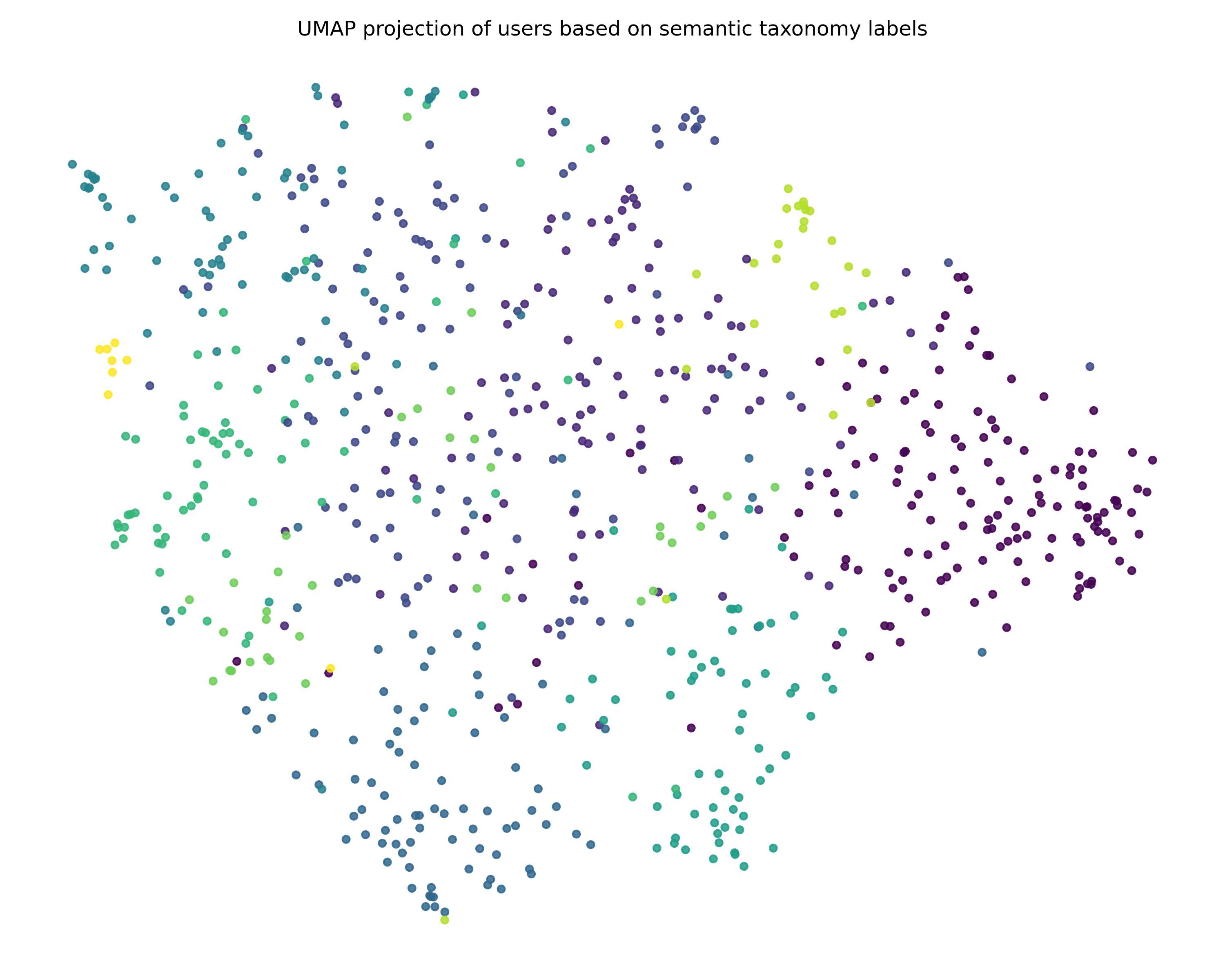}
        \caption*{(a) UMAP projection}
    \end{minipage}
    \hfill
    \begin{minipage}[t]{0.48\columnwidth}
        \centering
        \includegraphics[width=\linewidth]{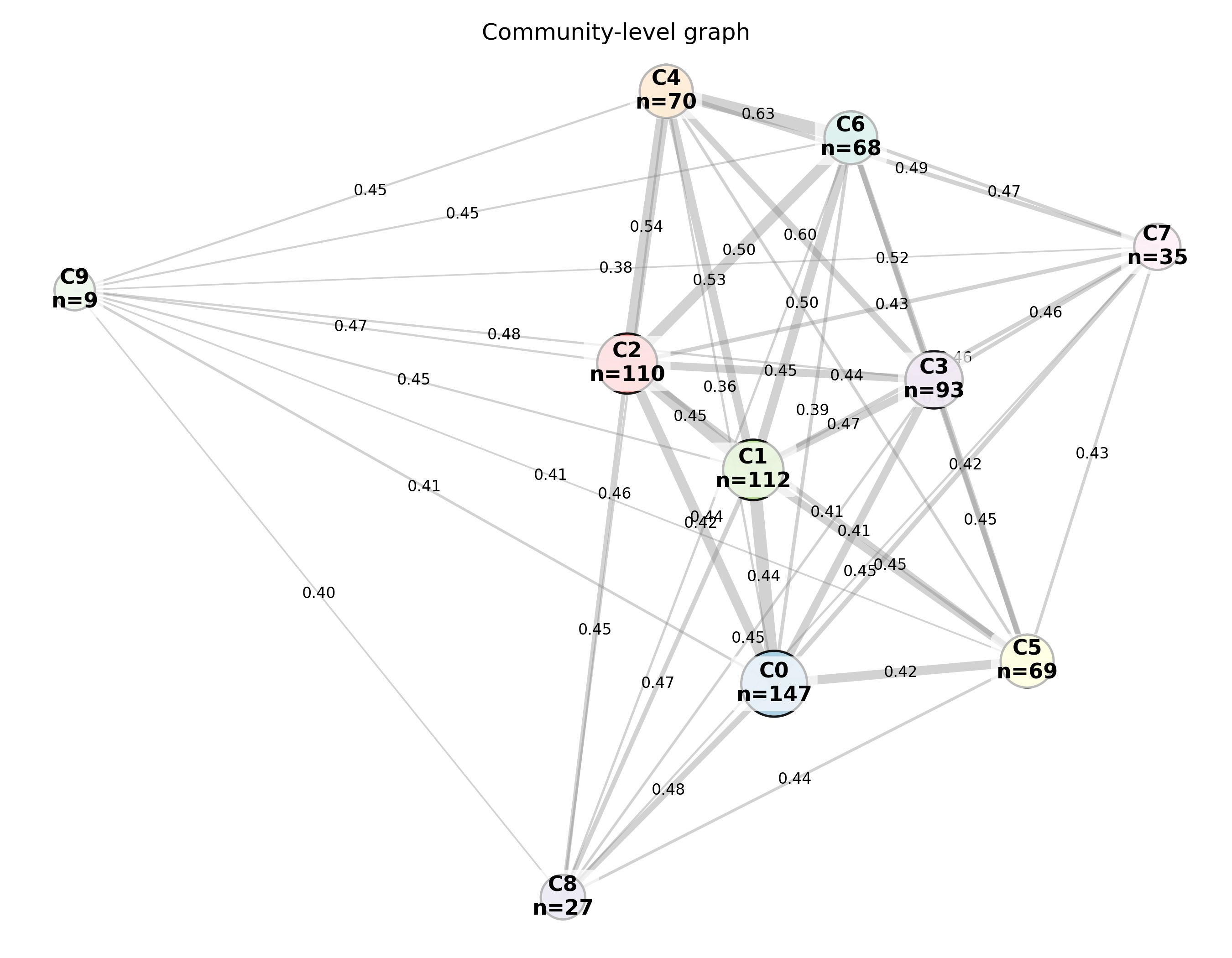}
        \caption*{(b) Community-level graph}
    \end{minipage}
    \caption{User communities based on semantic similarity.}
    \label{fig:semantic_umap_community}
\end{figure}

\begin{figure}[t]
    \centering
    \includegraphics[width=\columnwidth]{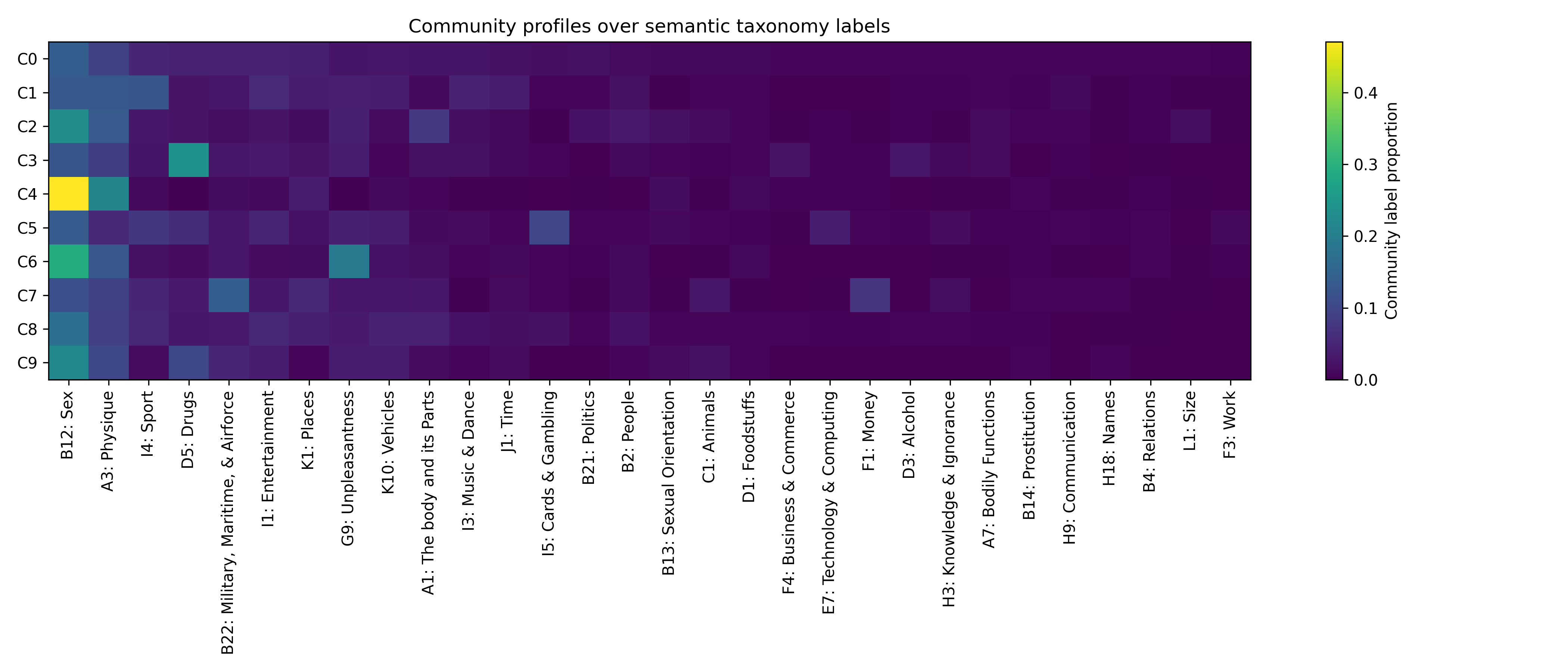}
    \caption{Semantic community distributions over labels.}
    \label{fig:semantic_heatmap}
\end{figure}

We construct a user similarity graph using semantic taxonomy labels mapped to the original \texttt{slang.gr} tags,
following the same process as above.
This yields a graph with 740 users and 11,544 edges, forming a sparse single connected component.
Leiden community detection 
identifies 10 communities with a weighted modularity of $0.44$,  indicating a clearer separation than in the tag-based graph.
Figure~\ref{fig:semantic_umap_community}(a) shows the UMAP projection, revealing partially separated regions that reflect broader topical affinities.
Figure~\ref{fig:semantic_umap_community}(b) shows that communities remain interconnected but are more differentiated than in the tag-based case, with a dense core and smaller, more weakly integrated communities (e.g., C8, C9) suggesting specialization.  Figure~\ref{fig:semantic_heatmap} reports the community distributions over the semantic labels, showing distinct profiles. 
C4 is dominated by sex and physique, C6 by unpleasantness and sex, C5 by gambling and technology, and C7 by military/maritime and money-related vocabulary. 
Community coherence is higher than in the tag-based graph, with the coherence lift ranging from $1.09$ to $1.93$ (mean $1.49$).

We replicate the user-level analysis on the semantic taxonomy graph, using the same thresholds. The distributions are shown in Table~\ref{tab:user_roles_comparison}.  Among the retained users, 17 are banned, compared to 21 in the tag-based graph. 
Banned users appear across roles, peaking among generalists (10.8\%). Peripheral users remain low (2.2\%), specialists none (0.0\%), while central actors such as core (5.4\%) and bridge users (2.7\%) show moderate presence.

\subsubsection{Metadata Taxonomy Similarity UC}

\begin{figure}[t]
    \centering
    \begin{minipage}[t]{0.48\columnwidth}
        \centering
        \includegraphics[width=\linewidth]{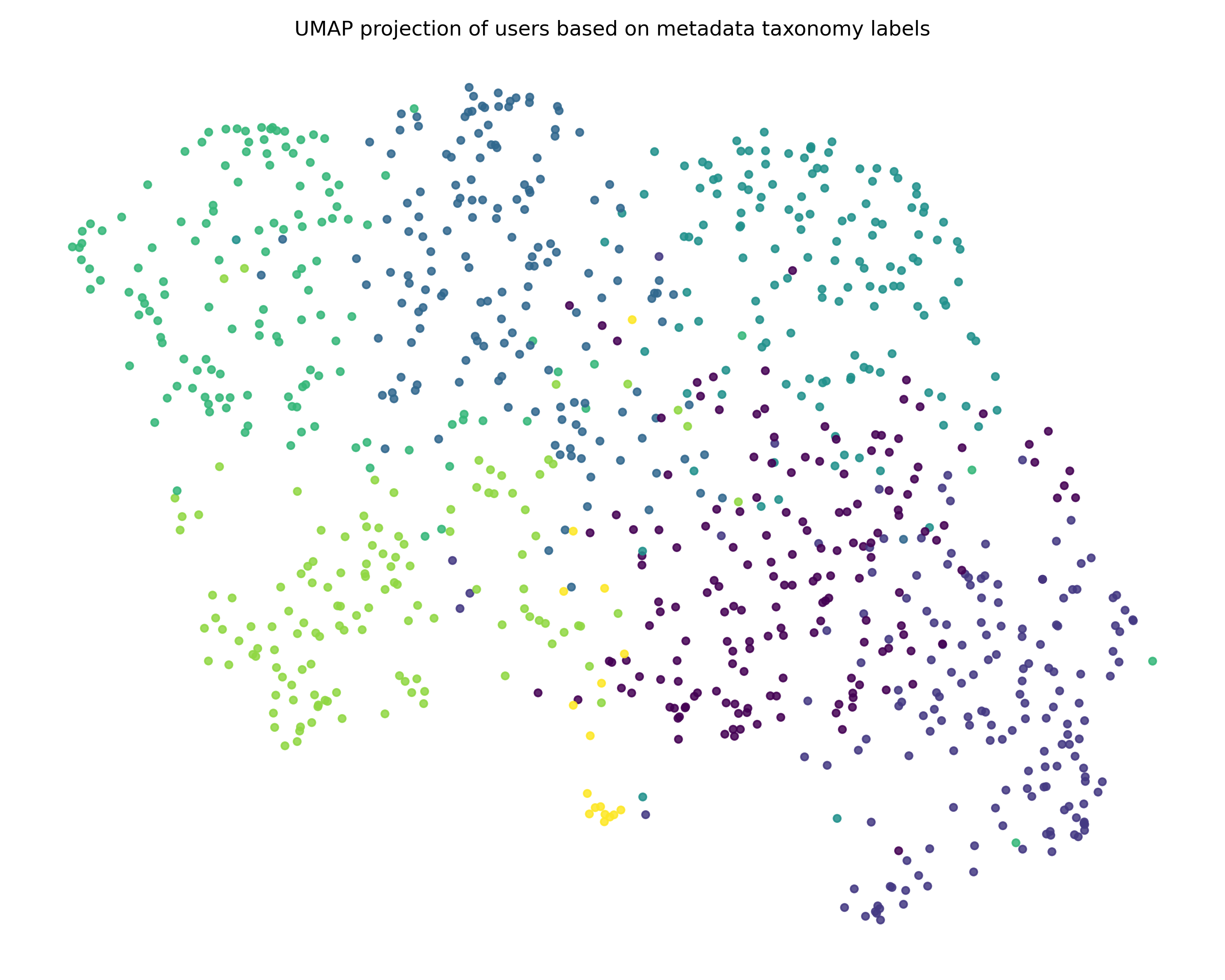}
        \caption*{(a) UMAP projection}
    \end{minipage}
    \hfill
    \begin{minipage}[t]{0.48\columnwidth}
        \centering
        \includegraphics[width=\linewidth]{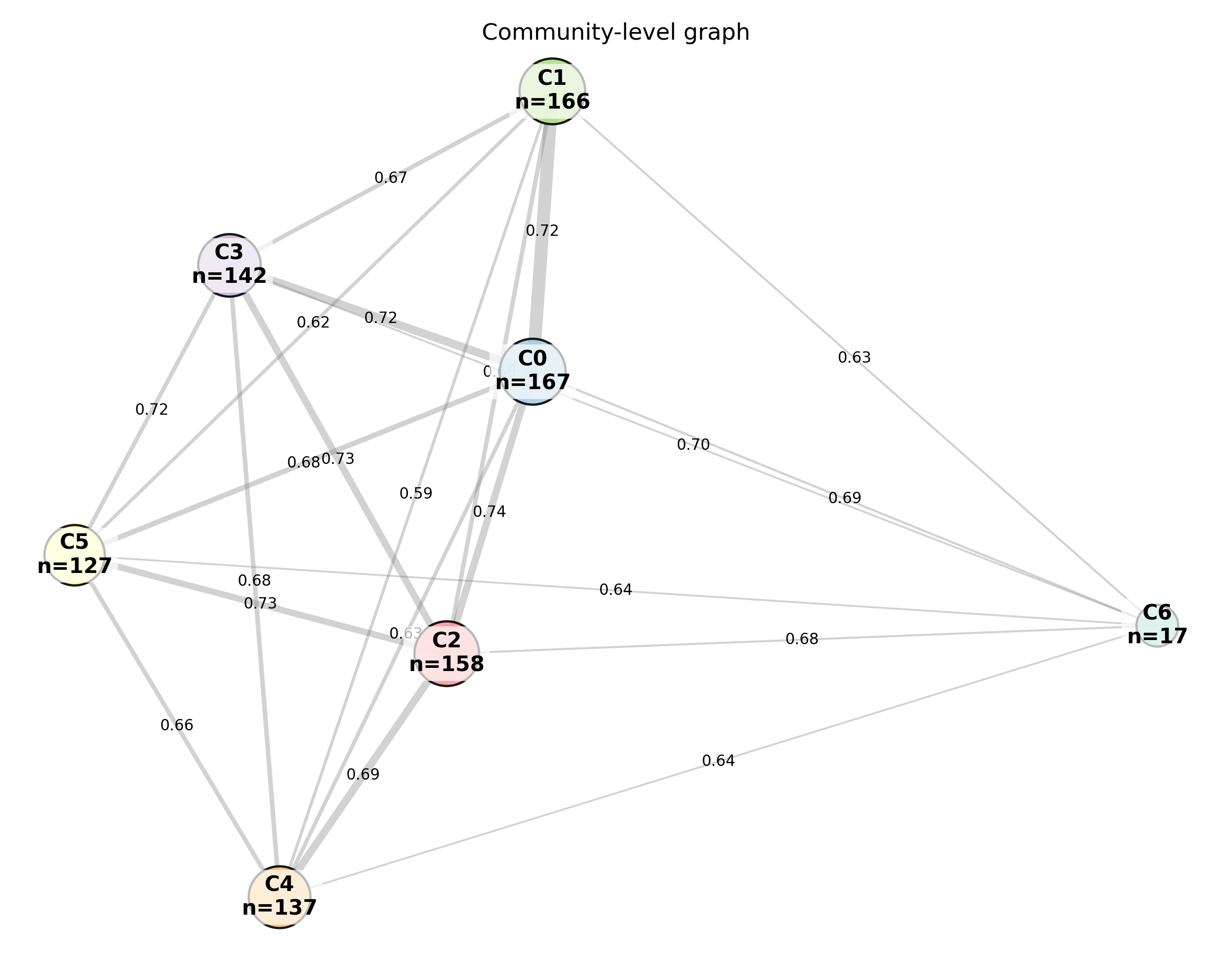}
        \caption*{(b) Community-level graph}
    \end{minipage}
    \caption{User communities based on metadata similarity.}
    \label{fig:metadata_umap_community}
\end{figure}

\begin{figure}[t]
    \centering
    \includegraphics[width=\columnwidth]{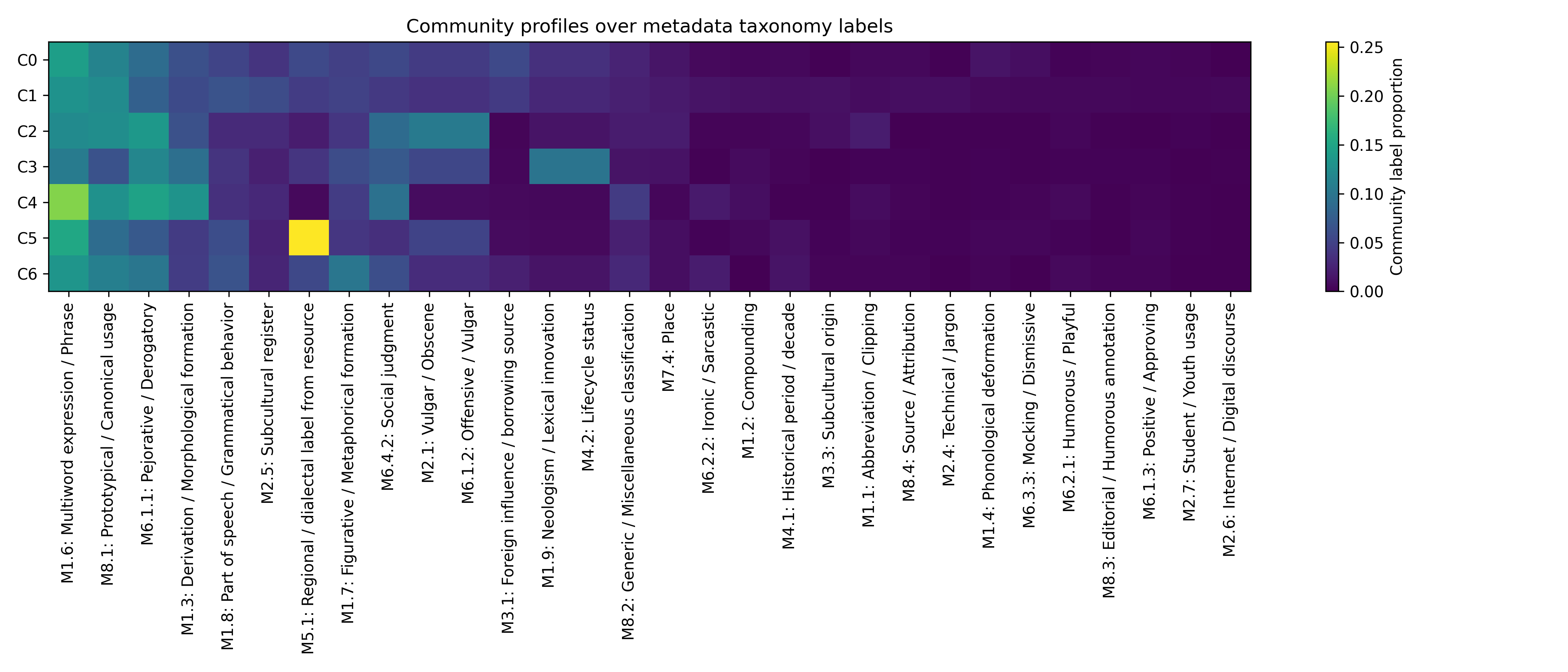}
    \caption{Metadata community distributions over labels.}
    \label{fig:metadata_heatmap}
\end{figure}

We construct a user similarity graph using the metadata taxonomy labels, following the same process.
The resulting graph has 914 users and 13,643 edges, forming a sparse single connected component.
Community detection identifies 9 communities with a weighted modularity of $0.48$, indicating clearer community separation under the metadata representation. 
The UMAP projection in Figure~\ref{fig:metadata_umap_community}(a) shows more visually distinct regions than the tag- and semantic-based projections. Together with the higher modularity, this suggests that metadata labels capture systematic differences in contributors’ stylistic, pragmatic, and sociolinguistic profiles. The community-level graph in Fig.~\ref{fig:metadata_umap_community}(b) shows that the communities nevertheless remain interconnected, with variation in edge strength indicating closer alignment among some groups and more peripheral positions for others. Figure~\ref{fig:metadata_heatmap} shows that the communities share a core profile characterized by frequent multiword expressions and pejorative or derogatory stance, while they differ selectively in canonical usage, derivational or morphological formation, and pragmatic features such as vulgar and offensive usage. Neologisms, regional or dialectal usage, and lexical innovation are more concentrated in particular communities, indicating localized specialization.
Although M7 Referent is globally frequent, it contributes relatively little to community differentiation because its most common labels, particularly person and state, are broadly distributed across users and communities. Their high document frequency results in lower TF–IDF weights, making them less discriminative than more unevenly distributed features from M6 Pragmatic stance and M2 Register. 
Community coherence is higher than in the tag-based graph, with coherence lift ranging from 0.99 to 1.59 (mean 1.33). 
Overall, the metadata communities share a broad common profile but display selective specialization in stylistic, pragmatic, regional, and morphological features. 

The user-level analysis is shown in Table~\ref{tab:user_roles_comparison}. 
The graph retains 914 users, showing a less aggressive filtering than the semantic graph. Among the retained users, 21 are banned, similarly to the tag-based case. 
Banned users are most frequent among generalists (8.7\%), followed by core (2.2\%) and bridge users (2.2\%). None of the peripheral users are banned, while the proportion among specialists remains low at 0.4\%.

\subsubsection{UC Discussion}
The three user-community graphs provide complementary views of contributor behavior rather than a single stable partition. The tag-based graph serves as a noisy baseline. It retains the fine-grained folksonomic variation, but it also produces more overlapping and less interpretable communities. The semantic taxonomy yields the most coherent communities, with the highest coherence lift, indicating that semantic normalization groups users with more similar topical profiles. By contrast, the metadata taxonomy achieves the highest modularity and retains more users, suggesting that stylistic, pragmatic, and sociolinguistic features provide a strong but complementary basis for community separation. Role distributions are broadly similar across graphs, with generalists consistently showing the highest proportion of banned users, while specialists and peripheral users show lower rates.  To assess consistency, we compare the partitions using ARI and NMI over the 682 users shared by all three graphs, confirming that the representations capture related but distinct aspects of behavior. Overall, agreement is low, with tag-based communities overlapping modestly with both semantic (ARI = 0.219, NMI = 0.192) and metadata partitions (ARI = 0.223, NMI = 0.215), while semantic and metadata partitions agree less (ARI = 0.156, NMI = 0.127), suggesting they capture more distinct aspects of user behavior. Thus, taxonomy-based representations do not simply reproduce the original tag structure but instead reorganize users along different dimensions of slang use. Semantic and metadata taxonomies provide complementary and only partially overlapping views of the \texttt{slang.gr} community.

\subsection{Comment Interaction \& Sentiment Graphs}

We analyzed comment sentiment using HellenicSentimentAI\footnote{A Greek RoBERTa classifier,  \url{https://huggingface.co/gsar78/HellenicSentimentAI}} mapping outputs to ($+1$, $-1$, $0$), with comments mostly neutral (61.5\%) and balanced positive (19.8\%) and negative (18.6\%) shares.
We construct directed positive ($IG^+$) and negative ($IG^-$) interaction graphs from commenters to definition authors. After filtering users with fewer than five comments, $IG^+$ has 1,323 nodes and 6,554 edges, and $IG^-$ has 1,228 nodes and 6,326 edges. Both are sparse and dominated by a giant component ($>99\%$ of nodes).  At the creator level, reception is measured via incoming positive and negative edges. We define \textit{controversial authors} as those in the top 25\% of both positive and negative reception (333 authors). Removing these authors from the top-10\% groups leaves only two authors with a strongly positive reception ($\geq 15$ positive comments) and none with a strongly negative reception. This indicates that most highly visible authors are not consistently positive or negative, but  receive both types of feedback. As a result, aggregated author-level sentiment alone is not a reliable indicator of socially accepted content. Banned users are active overall, but weakly represented in the sentiment graphs. Only a small subset appears as commenters or as authors receiving comments, and just one belongs to the controversial author set. Therefore, they provide little or no observable sentiment signal, suggesting that moderation is not well explained by aggregated author-level comment sentiment.
A similar pattern holds for commenters. We compute sentiment commitment scores by combining polarity with log-scaled comment volume and classify users into core positive,  negative, and controversial groups using the same thresholds. No banned users appear in any of these groups, including the 200 core controversial commenters. This suggests that moderation is not explained by aggregate comment sentiment alone, but likely depends on other behavioral or contextual factors. The undirected co-commenting graph connects users who comment on the same senses (429 users, 11,453 edges). Leiden community detection yields very low modularity ($Q = 0.080$, 6 communities), indicating overlapping interaction groups.  We also measure reciprocity, the mutual commenting between users, filtering users with less than five comments. Only 1,336 (14.8\%) of the 9,027 interacting user pairs exhibit reciprocal exchange.
Together, these results indicate that aggregated sentiment signals at both the author and commenter levels do not provide a reliable indicator of definition quality. Instead, sentiment primarily reflects engagement, definition editing, and visibility, while moderation and interaction structure are driven by broader behavioral and contextual factors.

\subsection{Community-Based Definition Confidence}

We assign a confidence score $C_g(d)$ to each definition $d \in D$ with author $u_d$ and comments $\mathcal{C}_d$, using user roles, comment sentiment, engagement, and moderation signals of a graph $g \in \{\mathrm{tag}, \mathrm{sem}, \mathrm{meta}\}$:
\[
C_g(d)=
\frac{
\alpha R^g(u_d)+
\beta S_d+
\gamma E_d+
\delta D^g_d-
\lambda B_d
}{
\alpha+\beta+\gamma+\delta
}.
\]
The resulting measure is a signed composite confidence index rather
than a probability bounded to \([0,1]\).

User reliability is defined from role membership:
\[
R^g(u)=
\begin{cases}
\dfrac{
\sum_{r \in \{\mathrm{core,spec,bridge,gen,per}\}}
w_r\,\mathbf{1}_r^g(u)
}{
\sum_{r} w_r
}, & \text{if } u \text{ has a role in } g,\\[1.2em]
w_{\mathrm{default}}, & \text{otherwise.}
\end{cases}
\]
where $\mathbf{1}_r^g(u)=1$ if user $u$ has role $r$ in graph $g$, $0$ otherwise, with
\[
(w_{\mathrm{core}},w_{\mathrm{spec}},w_{\mathrm{bridge}},
w_{\mathrm{gen}},w_{\mathrm{per}}, w_{\mathrm{default}})
=
(1.0,0.9,0.8,0.7,0.5,0.5).
\]
    
    Sentiment is computed per definition rather than per author:
    \[
    S_d =
    \begin{cases}
    \frac{1}{|\mathcal{C}_d|}\sum_{c \in \mathcal{C}_d} s(c) & |\mathcal{C}_d|>0\\
    0 & \text{otherwise}
    \end{cases},
    \quad s(c)\in\{-1,0,1\}.
    \]
    
    Engagement and diversity are:
    \[
    E_d =
    \frac{\log(1+|\mathcal{C}_d|)}{\max_{d'\in D} \log(1+|\mathcal{C}_{d'}|)},
    \quad
    D^g_d =
    \frac{-\sum_k p^g_{d,k}\log p^g_{d,k}}{\log K_g},
    \]
where $p^g_{d,k}$ is the fraction of commenters of $d$ in community $k$ of $g$, and $K_g$ is the communities number. If $|\mathcal{C}_d|=0$,  $E_d=D^g_d=0$.
    
    The moderation penalty is:
    \[
    B_d =
    \mathbf{1}[u_d \in \mathcal{B}]
    \]

    with $\mathcal{B}$ the set of banned users.
    
    We define two confidence scores:
    \[
    C_{\mathrm{tag}}(d), \quad
    C_{\mathrm{tax}}(d)
    =
    w \, C_{\mathrm{sem}}(d)
    +
    (1-w) \, C_{\mathrm{meta}}(d),
    \]
Using these settings ($\alpha=1.0$, $\beta=0.5$, $\gamma=0.2$, $\delta=0.2$, $\lambda=0.2$, 
role weights $(1.0,0.9,0.8,0.7,0.5)$,  $w_{\mathrm{default}}=0.5$, and semantic--metadata mixing weight $w=0.5$, the $C_{\mathrm{tag}}$ and $C_{\mathrm{tax}}$ scores exhibit similar but non-identical behavior. Across 28,384 senses, the mean $C_{\mathrm{tag}}$ is $0.301$ and the mean $C_{\mathrm{tax}}$ is $0.308$. The scores are moderately correlated (Spearman $\rho=0.673$), with $53.0\%$ overlap in the top $10\%$ (Jaccard $=0.360$), indicating substantial but incomplete agreement.
Ablation experiments show that user reliability is the main driver of divergence. Setting $\alpha=0$ yields identical rankings ($\rho=1.0$), while replacing role variation with a constant reliability ($w_{\mathrm{default}}$) makes the rankings highly similar ($\rho=0.927$, $93.9\%$ top-$10\%$ overlap). Removing sentiment reduces agreement ($\rho=0.614$, $43.8\%$ overlap), while engagement and diversity have smaller effects. Varying the taxonomy mixture shows that the semantic graph contributes more strongly to the upward shift in mean scores. Semantic-only scoring gives a mean confidence score of $0.319$ and $58.4\%$ top-$10\%$ overlap, compared to  $0.297$ and $46.9\%$ for metadata-only.
Overall, the taxonomy graph slightly shifts scores upward, especially through the semantic component, but ranking differences are mainly driven by user roles in each graph structure. The framework is fully parameterized, enabling systematic sensitivity analyses. The exploration of parameter variability and evaluation is left for future work.

\section{Conclusions and Future Work}
\label{sec:conc}

We presented the first large-scale computational study of \texttt{slang.gr} as a community-driven resource for Greek non-standard language, combining lexical, tagging, and interaction data to analyze its linguistic structure and contributor dynamics. Our analysis shows that Greek slang is strongly centered on person-related, embodied, and evaluative language, consistent with observations reported for  English, while showing morphological and pragmatic variation. At the community level, we identify highly skewed participation, short user lifespans, and moderately structured but overlapping communities. Across multiple representations, we observe consistent user roles, with broadly engaged users (generalists) more likely to be associated with moderation signals, linking diversity of participation with controversial or boundary-pushing content. We also define a confidence metric for definitions by combining user roles, interactions, and moderation signals. To support this analysis, we introduce a structured mapping of user-generated tags into semantic and metadata dimensions. While not a fully evaluated linguistic resource, this representation enables more interpretable and structured analyses compared to raw tags, and provides complementary views of topical and sociolinguistic variation. In the future, we plan to conduct extensive evaluations in downstream tasks. 

This work transforms \texttt{slang.gr} into a computationally usable resource for the study of non-standard Greek, supporting research on lexical variation, slang detection, generation, interpretation, and sociolinguistic NLP. The proposed taxonomy provides a first systematic framework for organizing Greek slang across semantic, sociolinguistic, pragmatic, and network-based definition confidence dimensions, and may facilitate cross-lingual alignment and multilingual analysis of informal language. It also enables the study of cases like bias and gender inequality, slang-based jailbreaking of LLMs, comparisons with standard Greek, and the analysis of semantic shift, linguistic creativity, and context-dependent meaning.

\bibliographystyle{ACM-Reference-Format}
\bibliography{references}
\end{document}